%% file: main.tex
\PassOptionsToPackage{dvipsnames,svgnames,x11names}{xcolor}
\documentclass[10pt,twocolumn,letterpaper]{article}

\usepackage[pagenumbers]{iccv} % To force page numbers, e.g. for an arXiv version

\input{preamble}

\definecolor{iccvblue}{rgb}{0.21,0.49,0.74}
\usepackage[pagebackref,breaklinks,colorlinks,allcolors=iccvblue]{hyperref}

\def\paperID{*****} % *** Enter the Paper ID here
\def\confName{ICCV}
\def\confYear{2025}

\title{Task-oriented Sequential Grounding and Navigation in 3D Scenes}

\author{%
Zhuofan Zhang$^{1,2}$\thanks{Work done as an intern at BIGAI.~~~\Letter~Corresponding author.}~~~Ziyu Zhu$^{1,2*}$ Junhao Li$^{1,2*}$ Pengxiang Li$^{1,3*}$ Tianxu Wang$^{1}$ \\
Tengyu Liu$^{1}$ Xiaojian Ma$^{1}$ Yixin Chen$^{1}$ Baoxiong Jia$^{1}$ Siyuan Huang$^{1}$ Qing Li$^{1\text{\Letter}}$ \\
 $^1$State Key Laboratory of General Artificial Intelligence, BIGAI, China\\
 $^2$Tsinghua University $^3$Beijing Institute of Technology \\
}

\begin{document}
\maketitle
\input{sec/0_abstract}    
\input{sec/1_intro}
\input{sec/2_relatedwork}
\input{sec/3_dataset}
\input{sec/4_approaches}
\input{sec/5_experiments}
\input{sec/6_conclusion}

{
    \small
    \bibliographystyle{ieeenat_fullname}
    \bibliography{reference_header, main}
}

\input{appendix}

\end{document}

%% file: preamble.tex
\usepackage{graphicx} \graphicspath{{figures/}}
\usepackage{amsmath}
\usepackage{colortbl} 
\usepackage[misc]{ifsym}
\usepackage[utf8]{inputenc} % allow utf-8 input
\usepackage[T1]{fontenc}    % use 8-bit T1 fonts
\usepackage{url}            % simple URL typesetting
\usepackage{booktabs}       % professional-quality tables
\usepackage{amsfonts}       % blackboard math symbols
\usepackage{nicefrac}       % compact symbols for 1/2, etc.
\usepackage{microtype}      % microtypography
\usepackage{subcaption}
\usepackage{caption}
\usepackage{array}
\usepackage{multirow}
\usepackage{graphicx}
\usepackage{tcolorbox}

\newcommand{\dataset}{SG3D\xspace}
\newcommand{\datasetnav}{SG3D-Nav\xspace}
\newcommand{\model}{SG-LLM\xspace}

\usepackage{pifont}

\definecolor{vblue}{HTML}{2993ba}
\newcommand{\blue}[1]{\textcolor{vblue}{\textbf{#1}}}

\newcommand{\blueprompt}[1]{\textcolor{blue}{ #1}}

%% file: sec/0_abstract.tex
\begin{abstract}
Grounding natural language in 3D environments is a critical step toward achieving robust 3D vision-language alignment. Current datasets and models for 3D visual grounding predominantly focus on identifying and localizing objects from static, object-centric descriptions. These approaches do not adequately address the dynamic and sequential nature of task-oriented scenarios. In this work, we introduce a novel task: Task-oriented Sequential Grounding and Navigation in 3D Scenes, where models must interpret step-by-step instructions for daily activities by either localizing a sequence of target objects in indoor scenes or navigating toward them within a 3D simulator. To facilitate this task, we present \dataset, a large-scale dataset comprising 22,346 tasks with 112,236 steps across 4,895 real-world 3D scenes. The dataset is constructed by combining RGB-D scans from various 3D scene datasets with an automated task generation pipeline, followed by human verification for quality assurance. We benchmark contemporary methods on \dataset, revealing the significant challenges in understanding task-oriented context across multiple steps. Furthermore, we propose \model, a state-of-the-art approach leveraging a stepwise grounding paradigm to tackle the sequential grounding task. Our findings underscore the need for further research to advance the development of more capable and context-aware embodied agents.

% that even state-of-the-art models struggle with sequential reliability in task-oriented scenarios. These findings highlight the need for further research in this area to improve performance on complex, sequentially grounded tasks.

% generated using GPT-4 with structured prompts and human-verified for quality. Benchmarking experiments show that state-of-the-art models struggle with this task, highlighting the need for further research for task-driven contexts.
\end{abstract}

\begin{figure*}[ht]
    \centering
    \includegraphics[width=0.8\linewidth]{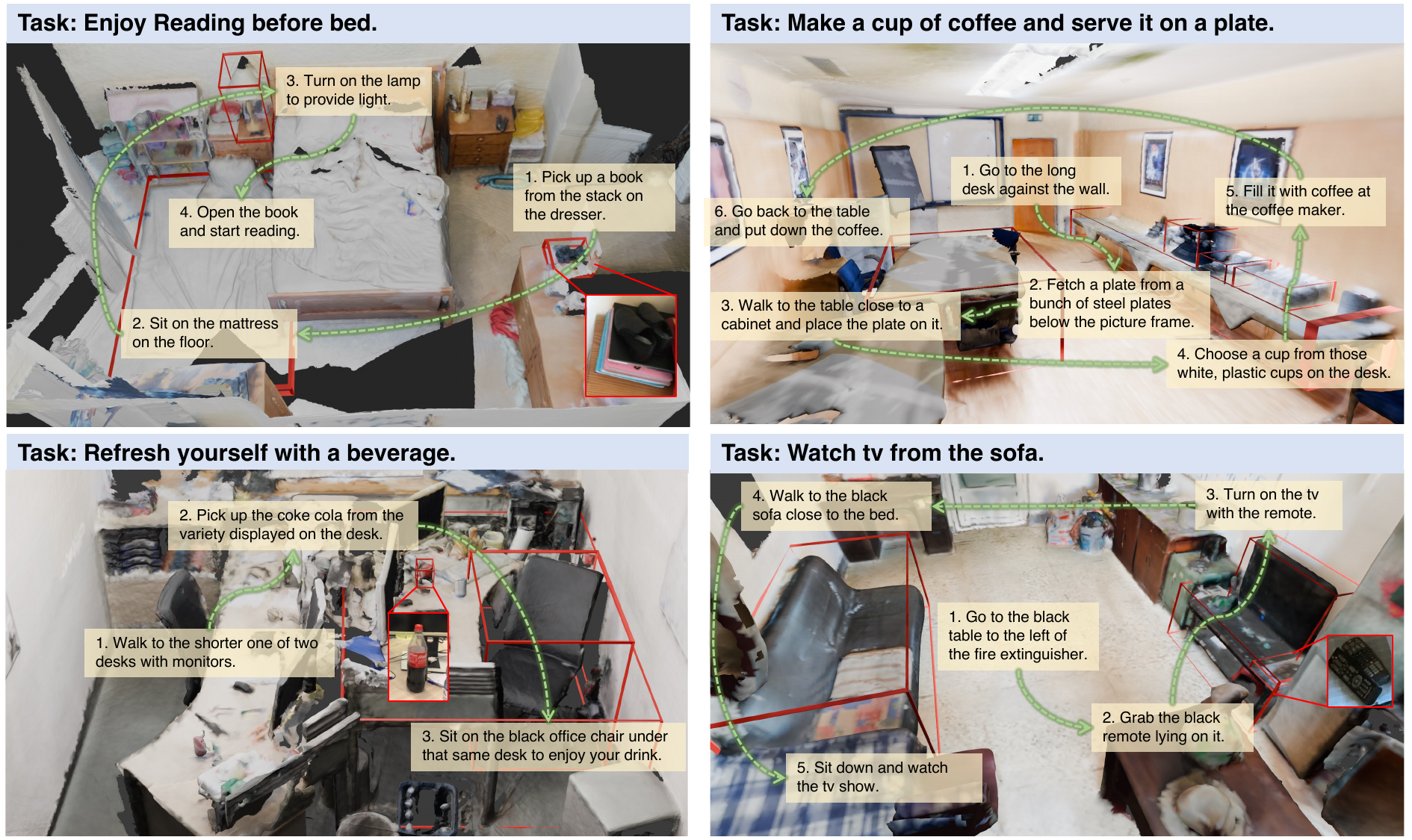}
    \caption{\textbf{The task-oriented \underline{s}equential \underline{g}rounding and navigation task in \underline{3D} scenes (\dataset)}, wherein models are required to interpret \textcolor{Goldenrod}{step-by-step} instructions for \textcolor{SteelBlue}{daily activities} by either localizing a sequence of \textcolor{BrickRed}{target objects} in indoor scenes or \textcolor{Green}{navigating toward} them within a 3D simulator. To solve this task, models must understand each step \textit{in the sequential context} to identify the target object, since a single step alone can be insufficient to distinguish the target from other objects of the same category.}
    \label{fig:teaser}
\end{figure*}

%% file: sec/1_intro.tex
\section{Introduction}
\label{sec:intro}

Grounding natural language in the 3D physical world is essential for aligning human and artificial intelligence in 3D world comprehension, a foundational capability for developing augmented/virtual reality systems~\citep{chen2022d,zhang2024towards} and embodied AI agents~\citep{embodied-survey,wang2023embodiedscan}. The 3D Visual Grounding (3D-VG) task addresses this challenge by identifying and localizing target objects through natural language descriptions within a 3D scene. Recent years have witnessed significant advancements in 3D-VG research, driven by the emergence of diverse datasets~\citep{jia2024sceneverse,scanrefer,referit3d,zhang2023multi3drefer,wang2023embodiedscan,kato2023arkitscenerefer} and innovative models~\citep{3d-vista,pq3d,vil3dref,guo2023viewrefer,butd}. Existing datasets primarily emphasize \textit{object-centric} descriptions~\citep{scanrefer,referit3d}, where target objects are identified through attributes and spatial relationships. However, this paradigm encounters limitations in practical scenarios where agents are expected to execute daily tasks, as exemplified in embodied AI studies like SayCan~\citep{saycan} and SayPlan~\citep{sayplan}, where high-level tasks are decomposed into \textit{sequential steps}. For each step, the agent must dynamically locate the relevant object in the scene to perform navigation or interaction. 
These cases' task-driven nature and multi-step structure are not involved in current 3D-VG studies.

% Thus, an overlooked gap exists between current 3D-VG studies and the demands of task-oriented scenarios.

% As shown in Embodied AI frameworks like SayCan~\citep{saycan} and SayPlan~\citep{sayplan}, high-level tasks are decomposed into sequential subtasks. For each subtask, the agent must dynamically locate context-dependent objects in the scene to perform navigation or interaction. This requirement exposes a critical gap: current 3D-VG methods struggle to handle task-driven object grounding that depends on procedural context rather than isolated object features.

% We further examine the gap between current 3D-VG datasets and the demands of task-driven scenarios. Firstly, the current 3D-VG datasets only involve single-step grounding, but in task-driven scenarios, a task plan usually involves multi-step. Secondly, step instructions have more focus on association on objects' affordence than object-centric descriptions.

We propose a novel task: task-oriented sequential grounding and navigation in 3D scenes. In this task, models are required to 
ground a step-by-step planned daily activity in an indoor 3D scene via identifying one target object for each step in the plan. The task consists of two settings: 1) Sequential grounding setting following 3D visual grounding task: The model localizes target objects within 3D scenes for each step. 2) Sequential navigation setting for embodied studies: The agent sequentially navigates to the target object of each step following the plan within a 3D simulator.

% \begin{enumerate}[itemsep=0pt, leftmargin=10pt]
%     \item Sequential Grounding Setting: The model localizes target objects within 3D point clouds for each step.
%     \item Sequential Navigation Setting: The agent sequentially navigates to the target objects to complete the plan.
% \end{enumerate}

\cref{fig:teaser} illustrates typical examples highlighting the task challenge. Beyond language style differences (object-centric versus task-oriented), models are expected to understand each step within the full \textit{context} of the task description and previous steps. For instance, in the ``Make coffee and serve it on a plate'' task (upper right), step 6 (``go back to the table and put down the coffee'') \textit{implicitly} references the same table from step 3. Distinguishing this table from the distractor table (at the bottom left of the view) requires contextual awareness that extends beyond the current step. Multi-step tasks often involve such latent object references across steps, which aligns with the human tendency to rely on context to interpret meaning in communication.

To facilitate this task, we construct a large-scale dataset named \dataset. We compile a set of RGB-D scans of realistic indoor scenes sourced from various 3D scene datasets, including ScanNet~\citep{scannet200}, ARKitScenes~\citep{baruch2021arkitscenes}, 3RScan~\citep{wald2019rio}, etc. These scenes encompass a variety of room types, such as bedrooms, kitchens, offices, bathrooms, and living rooms. We represent these scenes using 3D scene graphs~\citep{armeni20193d, wald2020learning} derived from SceneVerse~\citep{jia2024sceneverse}, which describe the objects' categories, attributes, and spatial relations within the scenes.

We further design an automated data-generation pipeline leveraging GPT-4~\citep{achiam2023gpt} to create diverse, high-quality daily tasks from these scene graphs. Each task comprises a high-level description and a step-by-step plan, with the target object annotated for each step. To ensure the validity of the generated tasks, we conduct a human verification process to check if the tasks were appropriate for the scenes, if the plans were sufficient to accomplish the tasks, and if the target objects were correctly identified for each step. Invalid tasks were either filtered out or manually refined. Ultimately, the proposed \dataset includes \textit{22,346 tasks} with \textit{112,236 steps} across \textit{4,895 real-world 3D scenes}. \cref{tab:stats_comparison} compares \dataset with existing 3D-VG datasets.
% Custom prompts encourage context-dependent task creation, while human verification filters illogical or hallucinated cases. Ultimately, \dataset includes \textit{22,346 tasks} with \textit{112,236 steps} across \textit{4,895 real-world 3D scenes}. 
% \cref{tab:stats_comparison} compares \dataset with existing 3D-VG datasets.

% To ensure the validity of the generated tasks, we conducted a human verification process to check if the tasks were appropriate for the scenes, if the plans were sufficient to accomplish the tasks, and if the target objects were correctly identified for each step. Invalid tasks were either filtered out or manually refined. Ultimately, the proposed \dataset includes \textit{22,346 tasks} with \textit{112,236 steps} across \textit{4,895 real-world 3D scenes}. \cref{tab:stats_comparison} compares \dataset with existing 3D-VG datasets.

% We further extend our task to a multi-object navigation setting~\citep{wani2020multion}. While the 3D visual grounding setting relies on global 3D point clouds as scene input, navigation tasks utilize ego-view RGB-D streams. In this setting, agents are not provided with the entire scene point cloud directly. Instead, they must acquire knowledge about the scene through active exploration. For this purpose, we use HabitatSim as the simulation environment and leverage \dataset data sourced from the HM3D dataset.

We introduce \model, a novel framework designed for sequential grounding by leveraging the reasoning and context-memorization capabilities of 3D-aware large language models (LLMs). Built on the Vicuna-7B backbone~\citep{chiang2023vicuna}, \model processes task descriptions and 3D object tokens to iteratively predict both procedural steps and their corresponding grounding targets. To ensure coherence across sequential steps, we propose a \textit{sequential adapter} that dynamically integrates embeddings of previously predicted objects into the LLM's attention mechanism. This enhancement enables context-aware object references, facilitating consistent and accurate grounding throughout the task.

We evaluate two experimental settings with separate baseline categories: 1) In the sequential grounding setting, we compare 3D visual grounding models~\citep{chang2024mikasa, guo2023viewrefer, 3d-vista, pq3d}, LLM methods~\citep{qwen2, qwen2.5, achiam2023gpt}, Chat-Scene~\citep{huang2024chat}, GPT4Scene~\citep{qi2025gpt4scene}, and the proposed \model model. \model outperformed all competing methods in the sequential grounding setting. 2) In the sequential navigation setting, we evaluate modular~\citep{fan2024embodied} and end-to-end architectures~\citep{khanna2024goat}. Our results indicate that task-oriented sequential grounding and navigation present significant challenges for current models, highlighting the need for further research and development in this domain.

% In our experiments, we adapted several state-of-the-art 3D visual grounding models to the sequential grounding task and evaluated their performance on the \dataset benchmark. The models included Vil3DRef~\citep{vil3dref}, MiKASA-3DVG~\citep{chang2024mikasa}, ViewRefer~\citep{guo2023viewrefer}, 3D-VisTA~\citep{3d-vista}, PQ3D~\citep{pq3d}, and LEO~\citep{leo}. Our results indicate that while these models perform well on previous benchmarks, they struggle with the more complex and realistic grounding scenarios presented in \dataset. Additionally, we adapted both an end-to-end model and a modular model to the sequential navigation task and evaluated them on the navigation data from \dataset. These findings highlight the need for further research and development to improve performance in task-oriented sequential grounding and navigation scenarios.

Our contributions are summarized as follows: 
\begin{itemize}[itemsep=0pt, leftmargin=10pt]
    \item We introduce a novel task, \textit{task-oriented sequential grounding and navigation}, extending 3D-VG research to task-driven scenarios.
    \item We construct a large-scale dataset for this novel task, \dataset, which contains 22,346 tasks with 112,236 steps across 4,895 real-world 3D scenes.
    \item We propose \model, a state-of-the-art approach leveraging stepwise grounding paradigm to tackle the sequential grounding task.
    \item We conduct a comprehensive benchmarking analysis on the grounding and navigation settings of \dataset separately. The results indicate that current models struggle with the task, highlighting the need for further advancements in this domain.
\end{itemize}

%% file: sec/2_relatedwork.tex
\section{Related Work}
\label{sec:relatedwork}

\begin{figure*}[t]
    \centering
    \includegraphics[width=0.8\linewidth]{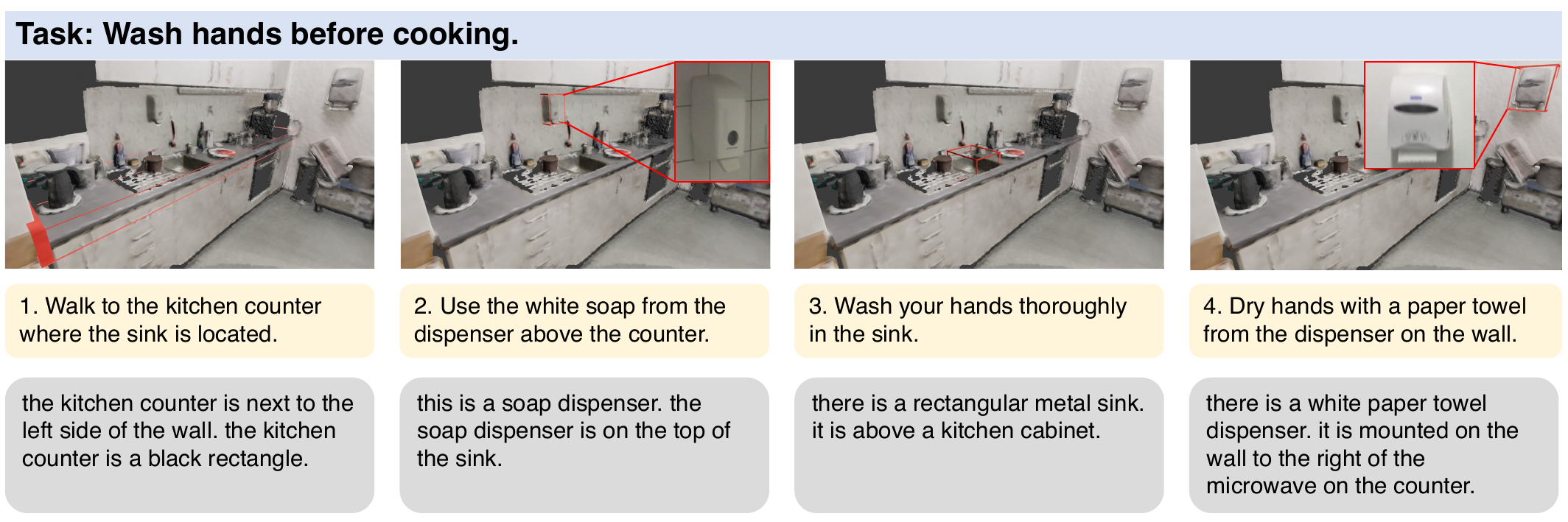}
    \caption{The comparison between task-oriented steps in \textcolor{Goldenrod}{\dataset (first row)} and object-centric referrals in \textcolor{Gray}{ScanRefer (second row)} for the same target objects. 
    % Particularly, in step 3, the ScanRefer annotation describes the sink's shape, material, and spatial relation to the cabinet to identify it, while the corresponding step in SG3D avoids such details. The \textit{context} provided by the task makes it easy to infer that the sink is near the soap dispenser mentioned in the previous step.
    }
    \label{fig:sg3d_vs_scanrefer}
\end{figure*}

\begin{table}[th]
    \small
    \centering
    \caption{\textbf{The comparison of \dataset with existing 3D visual grounding benchmarks.} \dataset expands the data scale of prior work by order of magnitude, especially in text length.}
    % ``VG'' stands for Visual Grounding, ``SG'' for Sequential Grounding, and ``MT'' for Multiple Tasks. * Only new data is counted.
    \label{tab:stats_comparison}
    \resizebox{\linewidth}{!}{
    \begin{tabular}{lcccccccccc}
    \toprule
    Dataset & Referral type & Scene & Obj. & Avg. Text Len. & Vocab. & Total \\
    \midrule
    ScanRefer~\citep{scanrefer} & Object-centric & 1.5K & 33K & 20.3 & 4,197 & 52K \\
    Nr3D~\citep{referit3d} & Object-centric & 1.5K & 33K & 11.5 & 2,986 & 42K \\
    Sr3D~\citep{referit3d} & Object-centric & 1.5K & 33K & 9.7 & 158 & 84K \\
    Multi3DRefer*~\citep{zhang2023multi3drefer} & Object-centric & 1.5K & 33K & 15.1 & 7,077 & 20K \\
    
    % SceneVerse*~\citep{jia2024sceneverse} & Object-centric & 68K & 1.5M & 14.7  & 24,304 & 2.2M \\
    \midrule
    \rowcolor[gray]{0.95} \blue{\dataset}  & Task-oriented & \blue{4.9K} & \blue{123K} & \blue{70.5} & \blue{8,136} & \blue{22K / 112K} \\
    \bottomrule
    \end{tabular}
    }
\end{table}

% \begin{table*}[th]
%     \small
%     \centering
%     \caption{\textbf{The comparison of \dataset with existing 3D visual grounding benchmarks.} \dataset expands the data scale of prior work by order of magnitude. ``VG'' stands for Visual Grounding, ``SG'' for Sequential Grounding, and ``MT'' for Multiple Tasks. * Only new data is counted.}
%     \label{tab:stats_comparison}
%     \resizebox{\linewidth}{!}{
%     \begin{tabular}{lcccccccccc}
%     \toprule
%     Dataset & Task & Referral type & Text Source & Quality Check & Scene & Obj. & Avg. Text Len. & Vocab. & Total \\
%     \midrule
%     ScanRefer~\citep{scanrefer} & VG & Object-centric & Human & \cmark & 1.5K & 33K & 20.3 & 4,197 & 52K \\
%     Nr3D~\citep{referit3d} & VG & Object-centric & Human & \cmark & 1.5K & 33K & 11.5 & 2,986 & 42K \\
%     Sr3D~\citep{referit3d} & VG & Object-centric & Template & \cmark & 1.5K & 33K & 9.7 & 158 & 84K \\
%     Multi3DRefer*~\citep{zhang2023multi3drefer} & VG & Object-centric & Template w/ Rephrasing & \cmark & 1.5K & 33K & 15.1 & 7,077 & 20K \\
%     %EmbodiedScan~\citep{wang2023embodiedscan} & VG & Object-centric & Template & \xmark & 5K & 890K &  &  & 970K \\
%     SceneVerse*~\citep{jia2024sceneverse} & MT & Object-centric & Human + GPT-3.5 & \cmark & 68K & 1.5M & 14.7  & 24,304 & 2.2M \\
%     \midrule
%     \rowcolor[gray]{0.95} \blue{\dataset} & SG & Task-oriented & GPT-4 & \cmark & \blue{4.9K} & \blue{123K} & \blue{70.5} & \blue{8,136} & \blue{22K / 112K} \\
%     \bottomrule
%     \end{tabular}
%     }
% \end{table*}

\noindent\textbf{3D Visual Grounding (3D-VG).}
3D vision-language learning seeks to connect natural language and the 3D physical world~\citep{3d-vista, pq3d, lerf}, enabling applications in augmented/virtual reality~\citep{chen2022d,zhang2024towards} and embodied AI~\citep{embodied-survey,wang2023embodiedscan}. This emerging field encompasses tasks such as 3D visual grounding (3D-VG, also referred to as 3D object localization), 3D question answering~\citep{scanqa, 3dgqa, sqa3d}, and 3D dense captioning~\citep{scan2cap}. Among these, 3D-VG---the task of identifying objects from candidate instances and localizing them via 3D bounding boxes in 3D scenes based on textual descriptions---has garnered significant attention. Recent advances include specialized benchmarks~\citep{scanrefer, referit3d, scanents, zhang2023multi3drefer, kato2023arkitscenerefer, zhu2024scanreason} and methods ranging from task-specific architectures~\citep{guo2023viewrefer, eda, 3dsps, butd, 3dvg, vil3dref, yang2024exploiting, shi2024aware, xu2024multi, lu2024scaneru, zhang2024cross, chang2024mikasa, llm-grounder, yuan2024visual} to unified 3D vision-language learning frameworks ~\citep{3d-vista, pq3d, unit3d}. Despite progress, existing 3D-VG benchmarks remain limited to object-centric, single-step grounding, overlooking the complexities of task-driven scenarios that demand multi-step grounding. \dataset provides more natural language and introduces diverse \textit{contextual} information.

\noindent\textbf{3D Large Language Models.}
Recent advancements in large language models (LLMs) have been significantly enhanced by integrating 3D spatial data, driving the development of 3D LLMs~\citep{3dllm-survey}. Models like 3D-LLM~\citep{3dllm} and Chat-Scene~\citep{huang2024chat} incorporate scene information via object-centric or point-level representations during instruction tuning~\citep{3dllm,pointllm,3DMIT,scene-llm,multiply}, while LL3DA~\citep{ll3da} enhances 3D perception using a Q-former-like architecture~\citep{blip2}. Recent works like LEO~\citep{leo}, 3D-VLA~\citep{3dvla}, and ManipLLM~\citep{manipllm} enable action-based interaction in 3D environments~\citep{leo,voxposer,moka}. Additionally, Chat-Scene~\citep{huang2024chat} frames 3D-VG as next-token prediction by associating each object instance in the scene with a textual identifier token. Building on these, we propose \model, a sequential grounding framework that harnesses the inherent reasoning power of 3D LLMs.
% Recent advancements in large language models (LLMs) have been significantly enhanced by integrating 3D spatial data, resulting in the development of 3D LLMs~\citep{3dllm-survey}. Existing works, such as 3D-LLM~\citep{3dllm} and Chat-Scene~\citep{huang2024chat}, use object-centric or point-level representations to incorporate scene information into LLMs during instruction tuning~\citep{3dllm, pointllm,3DMIT,scene-llm,multiply}. LL3DA~\citep{ll3da} employs a Q-former-like~\citep{blip2} structure to further improve LLMs' 3D scene perception. Concurrently, recent models like LEO~\citep{leo}, 3D-VLA~\citep{3dvla}, and ManipLLM~\citep{manipllm} have introduced action capabilities into 3D LLMs, enabling them to interact with and manipulate objects in 3D environments~\citep{leo, voxposer, moka}. Additionally, Chat-Scene~\citep{huang2024chat} enhances 3D-LLMs with grounding capability by associating each object instance in the scene with a textual identifier token, framing 3D-VG as a next-token prediction task. Building on these foundations, we propose \model, a sequential grounding framework that harnesses the inherent reasoning power of 3D LLMs. 
% However, \model attains grounding via specialized grounding tokens.
% Our work enhances the capabilities of 3D LLMs by incorporating grounding abilities, which output specific objects alongside the text.

\noindent\textbf{Embodied Navigation.} Embodied navigation tasks typically involve an agent operating within a simulated 3D environment, where the objective is to navigate to specific targets using egocentric camera observations~\citep{retrospectives}. In recent years, numerous benchmarks have been established for navigation tasks with varying target types, such as 3D coordinates, object categories, images, and object-centric language descriptions~\citep{dd-ppo, chaplot2020object, krantz2022instance, anderson2018vision}. ~\citet{wani2020multion} extends the object navigation task to a multi-object navigation task (MultiON), where the agent must navigate to an ordered sequence of target objects within a single episode. ~\citet{gireesh2023sequence} further generalizes MultiON by allowing sequence-agnostic targets, while GOAT~\citep{chang2023goatthing} and GOAT-Bench~\citep{khanna2024goat} extend MultiON to include multi-modal targets described through categories, images, or object-centric language. 
However, in these tasks, targets within a sequence remain independent. In contrast, targets within a sequence are usually interconnected in task-oriented sequential navigation, introducing a new layer of complexity.
% However, in these tasks, goals within a sequence remain independent, meaning the agent does not need to retain information about previous goals to locate subsequent ones. In contrast, our proposed task-oriented sequential navigation task requires the agent to navigate to a sequence of target objects based on a detailed plan of daily activities. Here, goals within a sequence are interconnected and specified by task-oriented context, introducing a new layer of complexity.

%% file: sec/3_dataset.tex
\section{3D Sequential Grounding and Navigation }

\subsection{Problem Formulation}
\textbf{Sequential Grounding} involves identifying temporally relevant objects for task completion. Formally, given a 3D scene $\mathcal{S}$ and a task $\mathcal{T}=(t,\{s_1,...,s_n\})$ where $t$ is a high-level task description and $s_1,...s_n$ are detailed steps of the task plan,
% stepwise subgoals
the model predicts a sequence of objects $\mathcal{O}=\{o_1,...,o_n\}$, \ie, learning a mapping $f: (\mathcal{S},\mathcal{T}) \to \mathcal{O}$, where each $o_i \in \mathcal{O}$ corresponds to the object grounded in step $s_i$.
% Compared to prior work, the challenge in our task lies in \textit{consistently} grounding objects across sequential steps of a task plan.

\noindent\textbf{Sequential Navigation} \noindent involves an agent sequentially navigating to the target object in a given task through a 3D simulator. Specifically, we employ a Stretch robot~\citep{khanna2024goat} with action space \{MOVE\_FORWARD (0.25m), 30º rotations, STOP\}. At each timestep, the agent processes: RGB-D observations $c_t, d_t$, pose data $(P_t;R_t)=(\delta_x,\delta_y;\delta_\theta)$, and current step $s_i$. The $i$-th step $s_i$ succeeds upon STOP action within 1m of the target object $o_i$ within 5000 timesteps from episode initialization.

% The sequential navigation setting involves an agent sequentially navigate to the target object in a given task. In specific, we follow ~\citet{khanna2024goat} to use HelloRobot's Stretch robot embodiment as the navigation agent, with the height of 1.41m and radius of 17cm. At each time step, the agent receives several sensory inputs including a 360 x 640 resolution RGB observation $c_t$, a corresponding depth image $d_t$, a GPS+Compass information $P_t=(\delta_x,\delta_y,\delta_z)$, $R_t=\delta_\theta$, as well as the current step instruction $a_i$. The agent's action space comprises MOVE\_FORWARD (by 0.25m), TURN\_LEFT and TURN\_RIGHT (by 30º), LOOK\_UP and LOOK\_DOWN(by 30º), and STOP actions. A sub-task in the episode is seen successful when the agent calls STOP action within 1m euclidean distance from the current goal object instance – within a budget of 500 agent actions (per sub-task).

\begin{figure}[!t]
    \centering
    % \vspace{-10mm}
    \includegraphics[width=\linewidth]{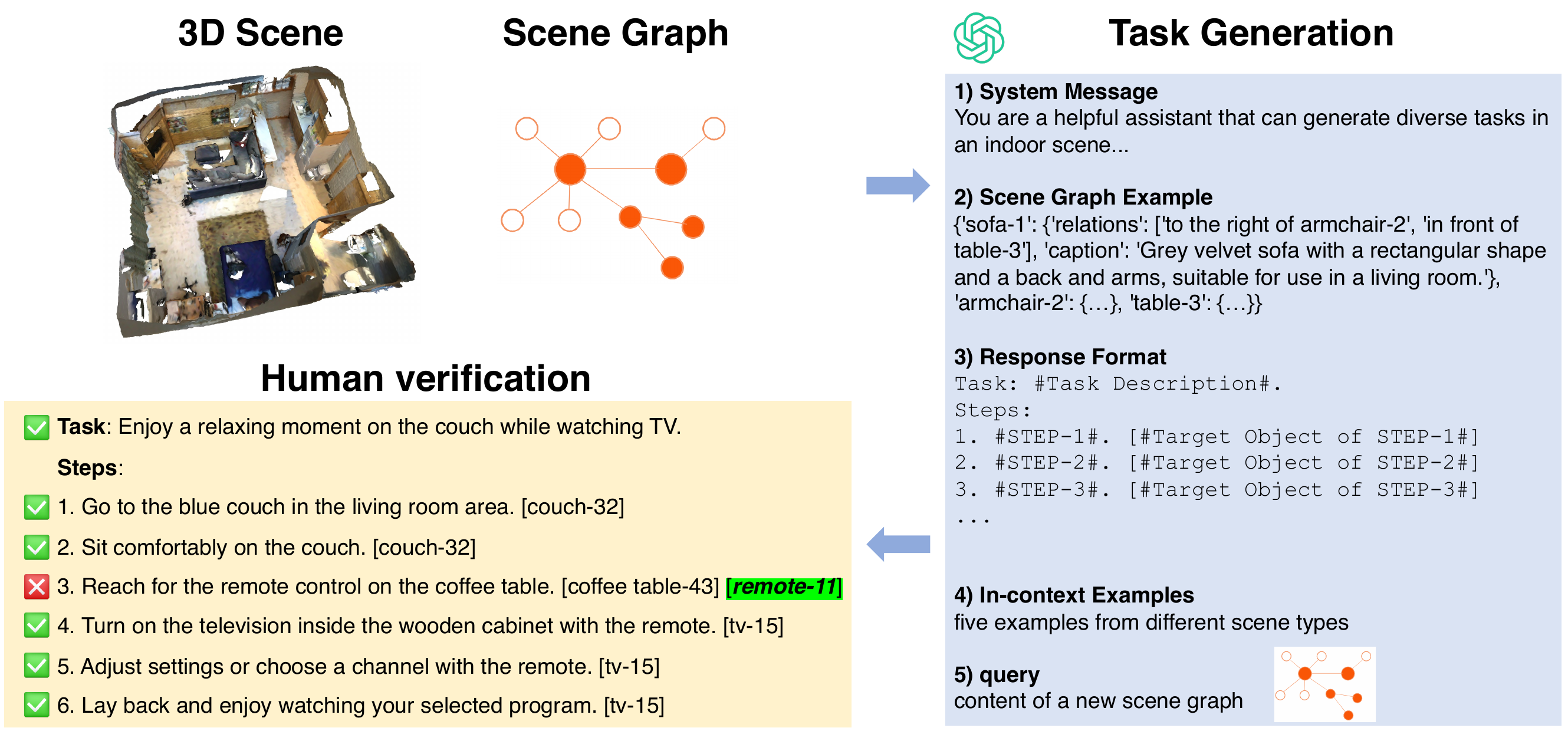}
    \caption{Data collection pipeline.}
    \label{fig:task_generation}
\end{figure}

\subsection{Dataset Construction}

As illustrated in \cref{fig:task_generation}, we leverage GPT-4 to generate tasks based on a 3D scene graph, followed by human verification.
% The full dataset is provided in the supplementary material.

\noindent\textbf{3D Scenes.} We select reconstructed scenes as the 3D environment for our task. Specifically, we utilized real-world scenes from the SceneVerse dataset, incorporating scenes from ScanNet, ARKitScenes, HM3D, 3RScan, and MultiScan. In total, we curate 4,895 3D scenes in \dataset. \cref{tab:scene_objects} presents the number of scenes used in each dataset and the average number of object instances per scene. We employ LLaMA-3~\citep{dubey2024llama} to analyze scene type distributions by processing object lists from each scene. The results indicate that most scenes correspond to five primary categories: bedrooms, offices, kitchens, bathrooms, and living rooms.
% Existing robotic task-planning approaches are typically evaluated in simulated environments~\citep{alfred,li2023behavior,sayplan}, lacking observation of their effectiveness in real-world scenarios. To address this, we select reconstructed scenes as the 3D environment for our task. Specifically, we utilized real-world scenes from the SceneVerse dataset, incorporating scenes from ScanNet, ARKitScenes, HM3D, 3RScan, and MultiScan. In total, we curate 4,895 3D scenes in \dataset. \cref{tab:scene_objects} presents the number of scenes used in each dataset and the average number of object instances per scene. We employ LLaMA-3~\citep{dubey2024llama} to analyze scene type distributions by processing object lists from each scene. The results indicate that most scenes correspond to five primary categories: bedrooms, offices, kitchens, bathrooms, and living rooms.

\noindent\textbf{Scene Graphs.} To provide GPT-4 with rich scene information, we process each scene into a semantic scene graph transformed from SceneVerse assets, which captures the categories, attributes, and spatial relations of objects within the scene. Each node in the graph represents a 3D object instance, while each edge represents a spatial relationship between nodes, such as ``near,'' ``below,'' or ``embedded.'' We further enhance these scene graphs by adding object captions provided in SceneVerse to enrich semantic details.

\noindent\textbf{Task Generation.} Using the 3D scene graph, we prompt GPT-4 to generate diverse tasks. We ask GPT-4 to create five distinct daily tasks for each scene. Each task comprises a general description $t$ and several steps $\{s_1,...,s_n\}$, with each step $s_i$ requiring the agent to focus on a specific target object $o_i$, such as navigating toward or interacting with it.

To balance diversity and coherence, we meticulously design prompts and provide in-context examples spanning five primary scene categories. These examples were intentionally crafted to emphasize contextual dependencies through strategic pronoun usage (e.g., ``it,'' ``here,'' ``the other'') while avoiding redundant descriptions. This methodology ensures generated tasks exhibit robustness, linguistic variety, and rich contextual relationships. For comprehensive documentation, the GPT-4 prompt and its analytical rationale are provided in \cref{sec:dataset construction details}.

% After generation, we remove any outputs with formatting errors and rigorously verify that all assigned targets are present in the corresponding scenes. Moreover, we observe that tasks exceeding ten steps tend to introduce hallucinated objects or problematic steps, which can negatively impact dataset quality. As a result, we discard any tasks containing more than ten steps.

\noindent\textbf{Human Verification.} We manually verify the evaluation set data to ensure quality. Given the 3D scene mesh and the generated task, annotators apply the following rules to judge each step's correctness: 1) If a step is deemed unfeasible or unrelated to the task description, it is marked as incorrect. 2) If a step is missing between step \( k \) and step \( k+1 \), step \( k+1 \) is judged as incorrect. 3) If the step’s description is insufficient to identify the target object, the step is considered correct only if the target object can still be inferred from the context; otherwise, it is marked as incorrect.
% \begin{enumerate}[itemsep=0pt, leftmargin=10pt]
%     \item If the step is unfeasible or unrelated to the task description, it is marked as incorrect.
%     \item If there is a missing step between step \( k \) and step \( k+1 \), step \( k+1 \) is judged as incorrect.
%     \item When the step's description is insufficient to identify the target object, the step is considered correct if the target object can still be identified through context; otherwise, it is marked as incorrect.
% \end{enumerate}
Tasks with a single incorrect step are manually revised, while those containing multiple incorrect steps are discarded. This rigorous human verification process ensures that the generated tasks are reasonable and the steps are feasible.
% A screenshot of the verification interface is provided under \cref{sec:dataset construction details}.
% As for steps like “Rub your hands” involving the agent itself rather than a specific object, we consider the target object from the previous step as the reference, implying "no need to change the position," which is reasonable in the navigation setting.

\noindent\textbf{\datasetnav.} We further introduce \datasetnav, a derivative navigation dataset developed using the Habitat-Sim simulator~\citep{savva2019habitat}. Derived from the HM3D split of \dataset, each task is converted into a navigation episode, requiring the agent to navigate to each step's target object sequentially. Details can be found in \cref{sec:dataset construction details}.
% Each step of the task is treated as a sub-task, with the target object of the step serving as the goal for that sub-task. 
% Each episode's starting position is randomly sampled such that their distance to the goal for the first sub-task lies between 1--30 meters. We filter episodes that involve goals exhibiting an IoU (Intersection over Union) score < 0.05 across all viewpoints within a 1-meter radius of its location.
% Finally, we collect 2,868 tasks with 12,385 steps across 181 real-world scenes in \datasetnav.

% Detailed statistics of SN3D are presented in FIGURE.
% The benchmark leverages real-world 3D scans from HM3DSem~\citep{hm3dsem}, which comprises 145 training and 36 validation house-level scenes. SN3D includes a total of XXX goal specifications and XXX episodes.

\begin{figure}
  \centering
  % \vspace{-10mm}
  \begin{subfigure}{0.49\linewidth}
    % \fbox{\rule{0pt}{2in} \rule{.9\linewidth}{0pt}}
    \includegraphics[width=\linewidth]{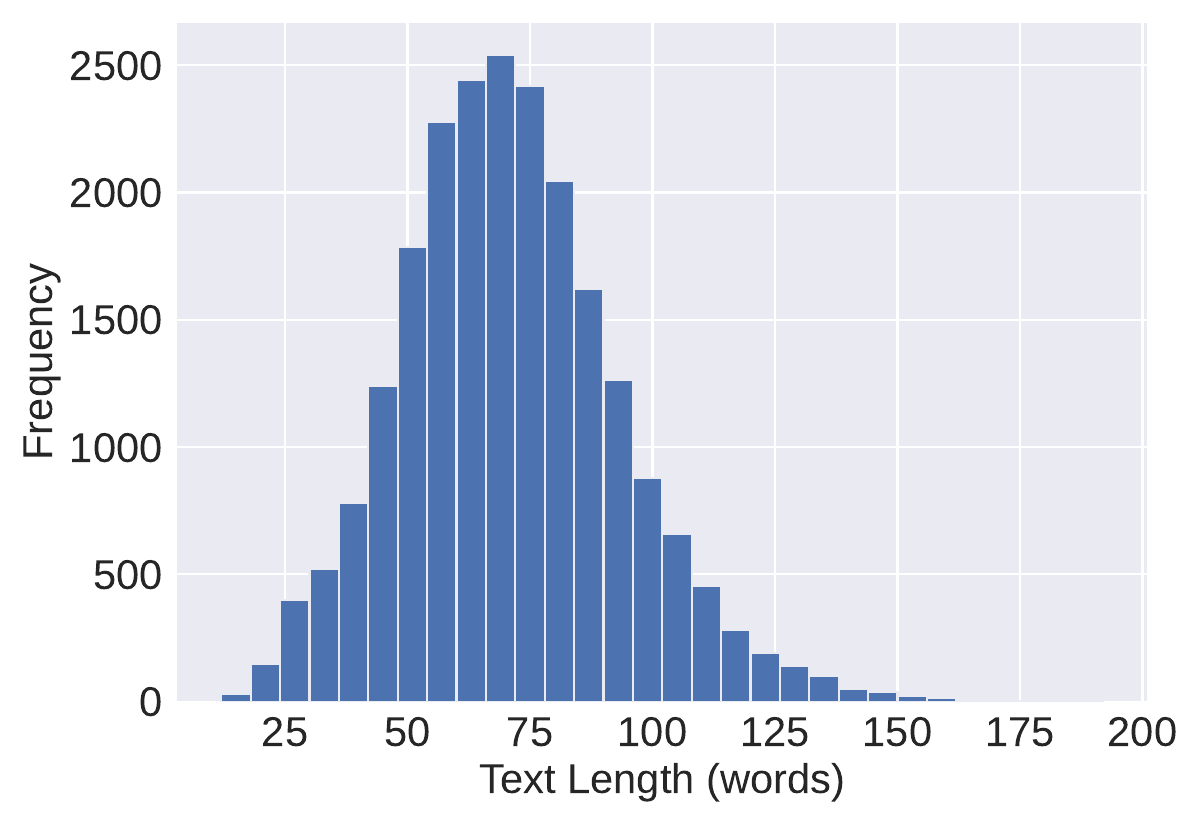}
    \caption{}
    % \caption{Step Counts.}
    \label{fig:step_counts}
  \end{subfigure}
  \hfill
  \begin{subfigure}{0.49\linewidth}
    % \fbox{\rule{0pt}{2in} \rule{.9\linewidth}{0pt}}
    \includegraphics[width=\linewidth]{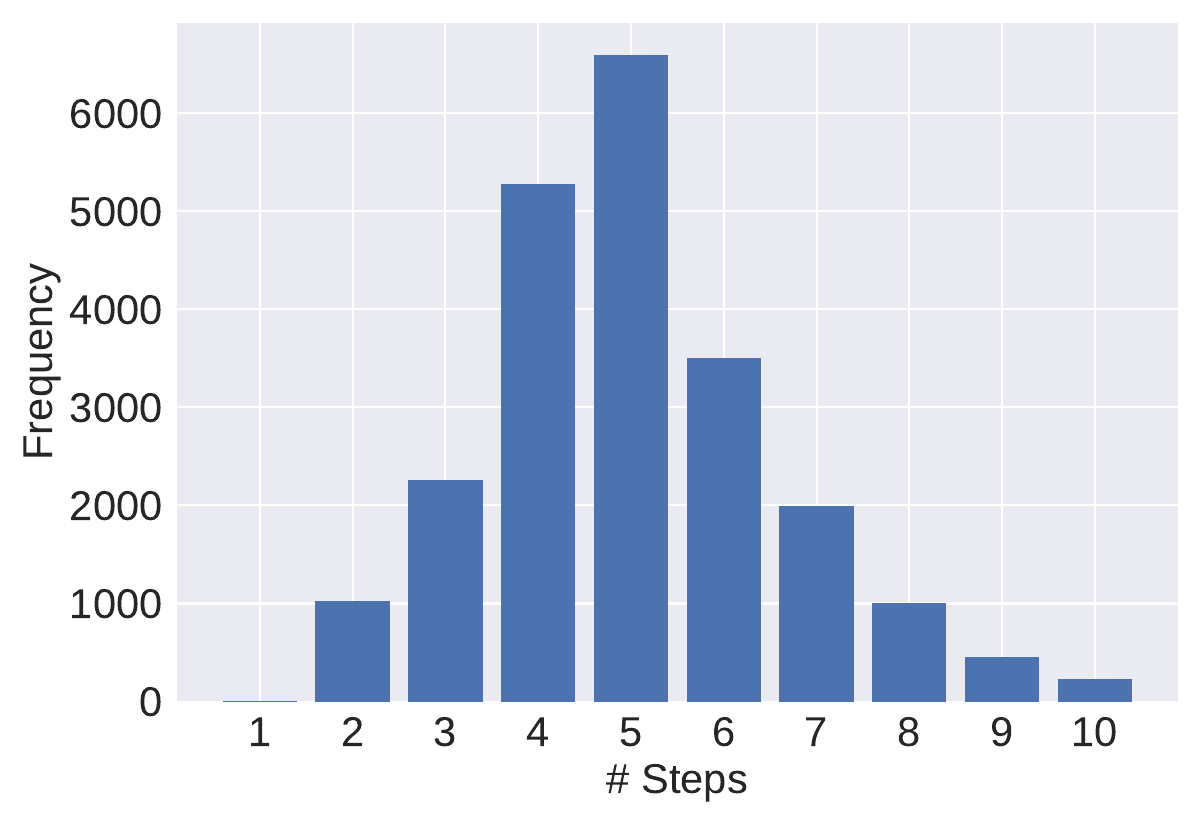}
    \caption{}
    % \caption{Text Length (words).}
    \label{fig:text_length}
  \end{subfigure}
  % \hfill
  % \begin{subfigure}{0.32\linewidth}
  %   % \fbox{\rule{0pt}{2in} \rule{.9\linewidth}{0pt}}
  %   \includegraphics[width=\linewidth]{target_id_counts.pdf}
  %   \caption{}
  %   % \caption{Target Object Counts.}
  %   \label{fig:target_id_counts}
  % \end{subfigure}
  \caption{Distributions of (a) text length (by words) per task, and (b) the number of steps per task.}
  \label{fig:distributions}
\end{figure}

\subsection{Dataset Analysis}

\begin{table}[t]
\centering
\caption{Dataset statistics of \dataset and \datasetnav.}
\label{tab:scene_objects}
\footnotesize
\begin{tabular}{lcccc}
\toprule
Dataset & \#scenes & \#obj. / scene & \#tasks & \#steps \\ 
\midrule
3RScan~\citep{wald2019rio} & 472 & 31.5 & 2,194 & 11,318 \\ 
ScanNet~\citep{dai2017scannet} & 693 & 30.7 & 3,174 & 15,742 \\ 
MultiScan~\citep{mao2022multiscan} & 117 & 40.8 & 547 & 2,683 \\ 
ARKitScenes~\citep{baruch2021arkitscenes} & 1,575 & 12.1 & 7,395 & 39,887 \\ 
HM3D~\citep{hm3d} & 2,038 & 31.0 & 9,036 & 42,706 \\ 
\midrule
\textbf{\dataset} & 4,895 & 25.1 & 22,346 & 112,336 \\ 
% \midrule
\rowcolor[gray]{0.9} \textbf{\datasetnav} & 181 & 664.2 & 2,868 & 12,385 \\
\bottomrule
\end{tabular}
\end{table}

In total, we collected data containing 22,346 tasks, encompassing 112,236 steps in \dataset. \cref{tab:scene_objects} presents the statistics of task and step counts in \dataset. Each task description has an average length of 6.9 words, and each step has an average length of 12.7 words.
% The dataset was split into training and evaluation sets. For 3RScan, scenes from its training and evaluation splits were used as our training set, while scenes from its test split were used as the evaluation set. For other datasets, we adhered to the original split of the 3D scenes provided.
\cref{fig:step_counts} illustrates the distribution of the number of steps per task, revealing an average of 5.03 steps per task. This underscores our benchmark's complexity and the data's sequential nature. \cref{fig:text_length} presents a histogram displaying the distribution of total text lengths for each task, including the task description and all steps, with an average of 70.5 words. This lengthy context poses a significant challenge for many text encoders, such as CLIP~\citep{radford2021learning}, indicating the need for models capable of handling lengthy inputs.

%% file: sec/4_approaches.tex
\section{Sequential Grounding Methods}

\begin{figure*}[t]
    \centering
    \includegraphics[width=\linewidth]{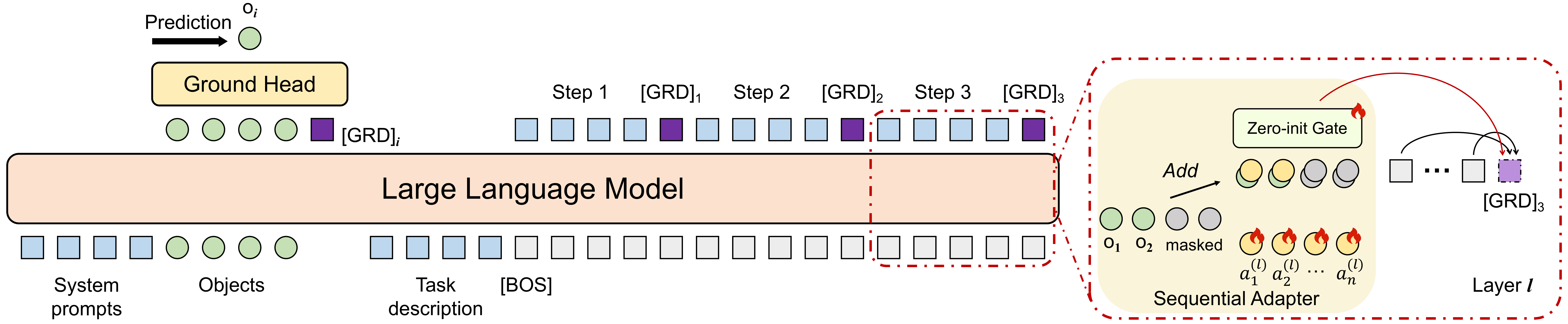}
    \caption{The structure of \model.}
    \label{fig:model}
\end{figure*}

We evaluate the following grounding approaches on the \dataset benchmark: 3D-VG baselines, LLM methods, 3D LLM baseline, large vision-laguage model baseline, and our proposed sequential grounding model \model. We follow ReferIt3D~\citep{referit3d} to decouple detection and grounding by using ground-truth object masks for all approaches.

% We explore several representative approaches for this purpose: three 3D-VL models depicted in \cref{fig:model}---the dual-stream model 3D-VisTA~\citep{3d-vista}, the query-based model PQ3D~\citep{pq3d}, the 3D LLM LEO~\citep{leo}. Additionally, we investigate the integration of GPT-4 with an object labeler. Further details are provided in the subsequent discussion.

\subsection{Baselines}

% We follow ReferIt3D~\citep{referit3d} to use ground-truth object masks. To ensure a fair comparison, we employ the point cloud as the scene representation and the same PointNet++~\citep{pointnet++} encoder to extract scene features for all three 3D-VL models.

\textbf{3D-VG Models.} We evaluate several state-of-the-art 3D-VG models on the sequential grounding task: MiKASA-3DVG~\citep{chang2024mikasa}, ViewRefer~\citep{guo2023viewrefer}, 3D-VisTA~\citep{3d-vista} and PQ3D~\citep{pq3d}.
% For a fair comparison, we use point clouds as the scene representation and employ the same PointNet++~\citep{pointnet++} encoder to extract scene features across all 3D-VL baselines.
Since none of the 3D-VG models inherently support multi-step grounding, each step in the plan requires an independent forward pass. Specifically, for step $i$, the input consists of the scene $S$ and the text, including the task description $t$, all prior steps $\{s_1, \dots, s_{i-1}\}$, and the current step $s_i$, with the output being $o_i$.

% \paragraph{3D-VL models} We also evaluated the performance of traditional 3D Vision-Language (3D-VL) models on the sequential grounding task. Our 3D-VL baselines include five state-of-the-art 3D-VL models: three specialized 3D visual grounding models---MiKASA-3DVG~\citep{chang2024mikasa}, ViewRefer~\citep{guo2023viewrefer}, and Vil3DRef~\citep{vil3dref}---and two unified 3D-VL models, 3D-VisTA~\citep{3d-vista} and PQ3D~\citep{pq3d}, which are capable of handling diverse 3D-VL tasks such as 3D grounding, question answering, and captioning. For a fair comparison, we use point clouds as the scene representation and employ the same PointNet++~\citep{pointnet++} encoder to extract scene features across all 3D-VL baseline models. Since none of these models inherently support multi-step grounding, each step of the plan requires an independent forward pass. Specifically, for step $i$, the input consists of the scene $S$ and the text, including the task description $t$, all prior steps $\{s_1, \dots, s_{i-1}\}$, and the current step $s_i$, with the output being $o_i$.

\noindent\textbf{LLMs.} Large Language Models (LLMs) have great potential for addressing multi-step contexts in the sequential grounding task, owing to their proficiency in long-text comprehension and logical reasoning. We assess the applicability of LLMs for sequential grounding by integrating them with a pre-trained PointNet++~\citep{pointnet++} classifier that predicts semantic categories for object instances in the scene. The LLM receives structured scene information in JSON format, including each object's ID, predicted category, and bounding box coordinates. The LLM generates a list of object IDs based on the scene information for each task.
% The specific prompt used is detailed in \cref{table::prompt_for_gpt_baseline}.
We evaluate four LLMs: GPT-4o, Qwen2.5-72b-Instruct~\citep{qwen2.5}, DeepSeek V3~\citep{liu2024deepseek}, and DeepSeek R1~\citep{deepseekai2025deepseekr1incentivizingreasoningcapability}. 
% We evaluated three LLMs: GPT-4o, Qwen2-72b-Instruct~\citep{qwen2}, and Qwen2.5-72b-Instruct~\citep{qwen2.5}. 
% Additionally, we tested these LLMs with ground-truth semantic categories to analyze their upper-bound performance. (this line to exp part)

\noindent\textbf{3D LLM.} In contrast to 3D-VG baselines, 3D LLMs offer enhanced flexibility in input-output formats and advanced reasoning capabilities. We adopt Chat-Scene~\citep{huang2024chat} as our 3D LLM baseline, a recent model that integrates visual grounding capacity into its 3D LLM architecture.

% \paragraph{3D LLM Baseline} We adopt Chat-Scene~\citep{huang2024chat} for its grounding capability in 3D object localization. The method assigns unique ID tokens to object instances by associating them with 3D feature tokens through a next-token prediction framework. Experiments are conducted on the \dataset ScanNet split, as the official implementation lacks cross-dataset preprocessing for other 3D benchmarks. We adapt the original grounding prompt structure for consistency.

\noindent\textbf{Large Vision-Language Model.} Recently, GPT4Scene~\citep{qi2025gpt4scene} advances 3D visual grounding benefiting from the development of Large Vision-Language Models (LVLM). Unlike traditional 3D-VG models and 3D LLMs that rely on point cloud features, GPT4Scene leverages RGB video frames to tackle visual grounding through an LVLM backbone.

\subsection{\model}

We propose \model for the sequential grounding task, leveraging the capabilities of 3D LLMs, specifically LEO~\citep{leo}, to address the task's inherent requirement for stepwise prediction.
% By framing the grounding task as a next-token prediction problem, 3D LLMs naturally align with our objective of reasoning over both previous and current step instructions.

\model's structure is depicted in \cref{fig:model}. We employ a spatial transformer to tokenize the 3D scene into a set of object tokens $\{o_j\}_{j=1}^N$, where each $o_j \in \mathbb{R}^{1 \times D}$ corresponds to the feature representation of the $j$-th object instance, and incorporate the task description $t$ as an input prompt for the LLM. We adopt Vicuna-7B~\citep{chiang2023vicuna} as the backbone LLM to process the object tokens and task description. As shown in \cref{fig:model}, \model processes the sequential grounding task through a stepwise paradigm: For each step $i$, \model is tasked to predict the step instruction $s_i$ along with a special grounding token $\texttt{[GRD]}_i$. This token is then concatenated with the object tokens and fed into a grounding head to predict the target object $o_i$ of step $i$. This design not only allows the model to effectively capture sequential dependencies but also equips it with task planning capabilities.

Inspired by \citep{zhang2023llama}, we propose a \textbf{sequential adapter}, an enhanced attention mechanism designed to enable stepwise grounding by explicitly incorporating the model's historical target object predictions. Specifically, our sequential adapter allows the model to attend to its own predicted target objects $o_1, \dots, o_{i-1}$ from prior steps when processing the current step instruction $s_i$ and its grounding token $\texttt{[GRD]}_i$.
% during the prediction of step instruction $s_i$ and the alongside grounding token $\texttt{[GRD]}_i$ at step $i$, extra attentions are allowed between the current token and the model's predicted target object for prior steps $o_1, \dots, o_{i-1}$, which are not directly available in LLM's autoregression. 
As illustrated in \cref{fig:model}, the adapter comprises learnable token embeddings $\{a^{(l)}\}_{l=1}^N$ across $N$ transformer layers, where $a^{(l)} \in \mathbb{R}^{K \times D}$ and $K$ correspond to the maximum number of steps observed in data. During step $i$, the adapter tokens at layer $l$ are dynamically augmented with the tokenized representations of previous target objects $o_1, \dots, o_{i-1}$, while future steps ($j \geq i$) are masked:
% \begin{equation}
%     \hat{a}_j^{(l)} = 
%     \begin{cases}
%         o_j + a_j^{(l)}, & \text{if } j < i, \\
%         0, & \text{otherwise}.
%     \end{cases}
% \end{equation}
\begin{equation}
    \hat{a}_j^{(l)} = \left(o_j + a_j^{(l)}\right) \cdot \mathbb{I}_{j < i},
\end{equation}
where $\mathbb{I}_{j < i}$ is an indicator function. The augmented adapter states $\hat{a}^{(l)}_1, \dots, \hat{a}^{(l)}_{i-1}$ are then stacked into a matrix $\hat{A}_{1:i-1}^{(l)} \in \mathbb{R}^{(i-1) \times D}$.
Let $M_i$ denote the token position of $\texttt{[GRD]}_i$ in the sequence. For each hidden state $h_M^{(l)}$ at layer $l$ corresponding to the $M$-th token $w_M$ ($M_{i-1} < M \leq M_i$), the adapter computes an auxiliary attention output over $\hat{A}_{1:i-1}^{(l)}$:
% To enhance the model’s ability to leverage historical grounding context, we introduce a supplementary attention mechanism that enables the model to explicitly reference object predictions from previous steps. During the generation of step $s_i$ and the alongside grounding token $\texttt{[GRD]}_i$ at step $i$, the model's query (for the $M$-th word token $t_{l}^{(M)}$ at layer $l$) attends to tokenized representations of previously predicted objects $o_1, \dots, o_{i-1}$.
% These object tokens $o_j \in \mathbb{R}^{1 \times D}$ are initially derived from a spatial transformer that encodes 3D scene inputs.
% Crucially, to address the semantic rigidity of reusing raw object tokens across transformer layers, we augment $o_j$ at layer $l$ with a learnable layer-specific vector $\Delta_j^l \in \mathbb{R}^{1 \times D}$, yielding:
% \begin{equation}
%     \hat{o}_j^{(l)} = o_j + \Delta_j^l.
% \end{equation}
% The attention operation then becomes:
\begin{equation}
    \begin{split}
        & \text{Attn}_{\text{adapter}}\left(h_M^{(l)}, \{\hat{a}_1^{(l)}, \dots, \hat{a}_{i-1}^{(l)}\}\right) = \\
        & \text{Softmax}\left(\frac{h_M^{(l)} W_Q^{l} \cdot \left(\hat{A}_{1:i-1}^{(l)} W_K^{l}\right)^\top}{\sqrt{D}}\right) \hat{A}_{1:i-1}^{(l)} W_V^{l},
    \end{split}
\end{equation}
This extra attention is combined with the transformer's native self-attention $\text{Attn}_{\text{vanilla}}^{(l)}$ through a learnable gating mechanism:
% with $\hat{A}_{1:i-1}^{(l)} \in \mathbb{R}^{(i-1) \times D}$ stacking the adapter tokens. This extra attention output $\text{Attn}_{\text{adapter}}$ is summated with the transformer’s vanilla self-attention output $\text{Attn}_{\text{vanilla}}^{(l)}$ via a gate $g^l$:
\begin{equation}
    \text{Attn}_{\text{total}}^{(l)} = \text{Attn}_{\text{vanilla}}^{(l)} + \tanh \left(G^l\right) \cdot \text{Attn}_{\text{adapter}}^{(l)},
\end{equation}
, where gate $G^l$ is initialized to zero to stabilize early training. By progressively integrating historical grounding decisions, the adapter fosters coherent object references across sequential steps.

During training, \model is optimized using two cross-entropy losses: A grounding loss aligns predicted object scores $f(\mathcal{S},\mathcal{T})$ with ground-truth indices $\mathcal{O}$, and an instruction loss supervises step generation $g(t, \mathcal{S})$:
% \begin{equation}
%     \mathcal{L}_{grd} = \mathbb{E}_{(\mathcal{S},\mathcal{T},\mathcal{O}) \sim \mathcal{D}} \text{CrossEntropy}(f(\mathcal{S},\mathcal{T}), \mathcal{O})
%     \label{eq:1}
% \end{equation}
\begin{equation}
\mathcal{L} = \mathbb{E}_{(\mathcal{S}, \mathcal{T}, \mathcal{O}) \sim \mathcal{D}} \Big[ \text{CE}\big(f(\mathcal{S}, \mathcal{T}), \mathcal{O}\big) + \text{CE}\big(g(t, \mathcal{S}), \mathcal{T}\big) \Big]
\label{eq:1}
\end{equation}
%During the inference phase, we employ two distinct settings to evaluate our models.
% During inference, the 3D-VL models receive the task description $t$ and detailed steps $\{s_1,...,s_i\}$, predicting the target object $o_i$ at each step $i$. For the \model, beam search with a beam width of 5 is employed to generate step instructions and the $\texttt{[GRD]}$ token.
% Specifically, at step $i$, the model can access the task description $t$, along with all prior and current steps $\{a_1,...,a_i\}$ while predicting the target object $O_i$. Dual-stream and query-based models, constrained by their architectures, require separate forward passes for each action step. In contrast, 3D LLM predicts target objects for all steps sequentially in a single forward pass.
% This setup applies to all models, facilitating a direct assessment of their ability to identify and prioritize the correct object based on sequential instructions.
%The second setting is specific to the 3D LLM and involves a more complex prediction task. In this scenario, the model is not only required to identify the target object but also to predict the sequence of steps leading up to that point. For this purpose, we utilize a beam search strategy to enhance the prediction accuracy and complexity management. The grounding head is activated specifically when a $\texttt{[GRD]}$ token is predicted, signaling the model to focus on grounding the relevant object in the context of the predicted sequence.

\section{Sequential Navigation Methods}
We benchmark two approaches on \datasetnav benchmark: a modular agent and an end-to-end policy.

\noindent\textbf{The Modular Agent.} Modular navigation approaches break down the navigation task into specialized, independent modules, each optimized separately before being integrated into a pipeline~\citep{chaplot2020object, chaplot2020neural, chaplot2020learning, chang2023goatthing}. Our modular baseline is derived from the multimodal Embodied VideoAgent~\citep{fan2024embodied} agent. During exploration, the agent constructs a dynamic object memory by extracting CLIP visual-language features \citep{radford2021learning} and estimating 3D positions for detected objects. At inference time, since modular architectures cannot maintain a memory of task context, the agent receives the task description $t$ and step-level instructions $\{s_1,\dots,s_i\}$ simultaneously. It then computes cosine similarity between the instruction’s CLIP text embedding and stored object features, selecting the highest-scoring object as the navigation target. The associated 3D position guides the agent toward the goal.
% During navigation, the agent matches the CLIP feature of the task description and step instructions $t,\{s_1,\dots,s_i\}$ (since the modular module cannot store task context, we input them in one go) with objects in memory using cosine similarity for text-image feature matching. The object with the highest similarity score is identified as the goal and localized to guide the agent's navigation.

\noindent\textbf{The End-to-End Policy.} In addition to evaluating the modular approach, we also benchmark an end-to-end navigation policy trained using reinforcement learning. Specifically, we adapt the monolithic sensor-to-act policy neural network from GOAT-Bench~\citep{khanna2024goat} for \datasetnav. The policy employs the CLIP~\citep{radford2021learning} encoder to generate embeddings for egocentric observations and step instructions $s_i$. These embeddings are then concatenated and processed through a GRU-based policy network to predict the agent's actions. The information provided by task-oriented context $t, \{s_1,\dots,s_{i-1}\}$ is implicitly stored in the GRU hidden state.

%% file: sec/5_experiments.tex
\section{Experiments and Results}
\label{sec:experiment}
\subsection{Evaluation Metrics}
% \paragraph{Evaluation Metrics}
We evaluate grounding performance using two metrics: step accuracy (s-acc) and task accuracy (t-acc). s-acc measures granular performance by averaging grounding correctness across all individual steps $s_i$. t-acc adopts a stricter holistic evaluation, where a task $t$ is deemed successful only if all constituent steps are correctly grounded. While s-acc quantifies basic object grounding capability at the step level, t-acc specifically tests the model’s capacity to consistently interpret and respond accurately across sequential steps.
% task accuracy (t-acc) and step accuracy (s-acc). Task accuracy refers to the average grounding accuracy over the total number of tasks $t$. A sample is considered correct if the grounded objects are accurately identified for all steps within a task. Conversely, step accuracy is calculated by averaging the accuracy across all individual steps $s_i$. Task accuracy evaluates the model's ability to consistently interpret and respond accurately across a sequence of text prompts. On the other hand, step accuracy focuses on the model's effectiveness at each individual step.
Correspondingly, we assess navigation performance via three metrics: step success rate (s-SR), task success rate (t-SR), and SPL~\citep{batra2020objectnav}. s-SR, aligned with s-acc, averages success rates across individual steps $s_i$. t-SR, analogous to t-acc, measures the proportion of fully completed tasks,
where success requires all steps in the task episode to be successful. To assess trajectory optimality, we employ the SPL metric with s-SR.
% For navigation assessment, we employ three metrics: task success rate (task-SR) and step success rate (step-SR). Task-SR, corresponding to t-acc, is the average navigation success rate over the total number of tasks $t$. A task is considered successful if all steps of the episode succeed. Step-SR, corresponding to s-acc, is the average success rate over all steps $s_i$. In addition, we employ a path efficiency metric SPL~\citep{batra2020objectnav} combining step-SR with path optimality.

% For the LEO model, we conduct a specialized evaluation to assess its capabilities in task planning and object grounding. Rather than providing LEO with pre-determined task plans, we challenge the model to autonomously generate both the plan and the corresponding grounded objects. To validate the accuracy and feasibility of the plans and grounding produced by LEO, we employ a GPT-based model. Following the approach used in OpenEQA, we utilize a prompt-based method with GPT to verify the effectiveness of the plans and groundings generated by LEO. This method allows us to critically assess whether LEO's outputs align with expected task outcomes and object interactions, ensuring that the model's autonomous planning and grounding are both logical and applicable to the given scenarios.

\subsection{Quantitative Results \& Analysis}

\subsubsection{Results on Sequential Grounding Benchmark}

\textbf{1. State-of-the-art 3D visual grounding models exhibit limited zero-shot transferability to \dataset.} As shown in \cref{tab:main-2}, 3D-VG models achieve subpar performance in zero-shot settings, with s-acc (14.2\%--34.6\%) and t-acc (0.0\%--11.7\%) across datasets. The discrepancy underscores \dataset's distinct challenges compared to conventional visual grounding benchmarks.
% This indicates that the models' pre-training on non-sequential tasks is insufficient for handling the complexities inherent in sequential grounding, highlighting the need for task-specific fine-tuning.

\noindent\textbf{2. LLM-based methods exhibit superior generalization capabilities over 3D-VG models on \dataset.
% but coarse scene perception hinders their performance.
} GPT-4o, Qwen2.5, and DeepSeek V3 achieve s-acc (30.0\%--33.3\%) and t-acc (10.3\%--11.7\%), slightly surpassing the zero-shot results of 3D-VG models. This edge stems from LLMs' advanced in-context learning and reasoning capabilities. Notably, the reasoning-focused LLM, DeepSeek R1, achieves significantly higher performance (40.2\% s-acc, 14.2\% t-acc), underscoring the critical role of reasoning in \dataset. However, the performance of LLMs is still low, constrained by their limited scene awareness. 3D LLM Chat-Scene also surpasses 3D-VG models for zero-shot inference, while its reliance on LLMs trained for non-sequential tasks limits effectiveness in sequential reasoning scenarios, causing it to underperform LLM methods. However, GPT4Scene, a VLM enhanced with fine-grained video features, outperforms both LLM and 3D-VG baselines, demonstrating the value of feature quality.
% \textbf{2. LLM-based methods demonstrate strong reasoning potential but suffer from 3D perception limitations on \dataset.} GPT-4o, Qwen2, and Qwen2.5 achieve s-acc (28.5\%–32.8\%) and t-acc (8.3\%–11.5\%), marginally outperforming 3D-VL models' zero-shot results.
% The performance hierarchy (Qwen2 < GPT-4o < Qwen2.5) reveals a clear scaling trend with increasing LLM capability.
% Notably, substituting PointNet++-predicted semantic categories with ground-truth labels yields dramatic improvements: GPT-4o's s-acc rises to 73.4\% (+43.4pp) and t-acc to 46.6\% (+36.3pp), underscoring both the critical role of 3D perception and the untapped potential of LLMs' reasoning abilities.

\noindent\textbf{3. Fine-tuning greatly enhances performance, but low task accuracy scores (< 40\%) indicate that consistent sequential grounding remains a challenge.} For example, 3D-VisTA's t-acc increases from 8.0\% to 29.5\%, while PQ3D's t-acc improves from 7.5\% to 25.7\%. \model achieves the best performance after fine-tuning, with an s-acc of 63.1\% and a t-acc of 33.9\%. Despite these improvements, their t-acc scores remain below 40\%, indicating that current models still face challenges in achieving consistent sequential grounding.

% \textbf{3. Fine-tuning greatly enhances performance but low task accuracy scores (< 40\%) indicate that consistent sequential grounding remains a challenge.} 3D-VisTA's t-acc increases from 8.3\% to 29.3\%, while PQ3D's t-acc improves from 7.8\% to 25.7\%. \model achieves the best performance after fine-tuning, with a s-acc of 63.0\% and a t-acc of 32.9\%. Despite these improvements, all models' t-acc scores remain below 40\%, indicating that current models still struggle to achieve consistent sequential grounding. This limitation highlights the need for further research and model design to effectively address the challenges posed by sequential grounding tasks. 

\noindent\textbf{4. \model consistently outperforms other baselines across all datasets, particularly in task accuracy.} \model achieves the highest s-acc 63.1\% and t-acc 33.9\% among all approaches. 
This superior performance stems from \model's 3D LLM architecture and memory mechanism, which effectively capture and reason about sequential dependencies in grounding tasks. While \model also enhances step accuracy, the gains are less pronounced compared to the substantial improvements observed in task accuracy.

% \textbf{4. The combination of GPT-4 and 3D object classifier is insufficient for addressing the sequential grounding task.} Despite GPT-4's robust reasoning and generalization capabilities, its performance---recording a t-acc of 7.6\% and a s-acc of 27.3\%---is significantly inferior to that of fine-tuned 3D vision-language models. This shortfall can be attributed to classification inaccuracies and the loss of information when translating the scene into semantic labels and positions. These results indicate that the effectiveness of large language models in this problem is heavily influenced by the alignment between 3D vision modality and text modality, making 3D-VL models the more suitable approach.

\subsubsection{Results on Sequential Navigation Benchmark}
Though sequential navigation presents challenges for both modular and end-to-end approaches, the end-to-end method demonstrates superior consistency in comprehending navigation targets. As shown in \cref{tab:nav_result}, both frameworks exhibit limited success across all metrics (s-SR: 12.1\%--14.7\%, t-SR: 3.8\%--7.7\%, SPL: 10.1\%--10.2\%), highlighting the inherent complexity of sequential navigation. The modular agent, while excelling in atomic navigation---surpassing the end-to-end policy in s-SR (+3.8\%) and SPL (+1.5\%)---fails to model inter-step dependencies, leading to critical shortcomings in context reasoning. In contrast, the end-to-end reinforcement learning policy achieves substantial gains after \datasetnav fine-tuning (+6.9\% s-SR, +6.5\% t-SR, +5.3\% SPL), ultimately outperforming the modular approach in t-SR by 3.9\%. This divergence suggests that modular architectures, though effective for isolated steps, lack mechanisms to retain context across steps without task-specific fine-tuning. End-to-end methods, however, inherently capture sequential logic through hidden states that encode prior targets, enabling more coherent navigation. 

\begin{table*}[t]
% \vspace{-5mm}
\centering
 \caption{\textbf{The grounding accuracy on \dataset.} ``s-acc'' denotes the grounding accuracy averaged over steps and ``t-acc'' denotes the grounding accuracy averaged over tasks. A task is considered correct if and only if all steps are correct. 
 % We ran each experiment three times and reported error bars.
 }
    \label{tab:main-2}
\resizebox{1\textwidth}{!}{
\begin{tabular}{llcccccccccccc}
\toprule & \multirow{2}{*}{Model Type} & \multicolumn{2}{c}{ScanNet} & \multicolumn{2}{c}{3RScan} & \multicolumn{2}{c}{MultiScan} & \multicolumn{2}{c}{ARKitScenes} & \multicolumn{2}{c}{HM3D} & \multicolumn{2}{c}{Overall} \\
&   & s-acc  & t-acc  & s-acc & t-acc & s-acc & t-acc & s-acc  & t-acc  & s-acc & t-acc & s-acc & t-acc \\ \midrule
\rowcolor[gray]{0.9} \textbf{Zero-shot} & & & & & &  & & & & & & & \\
MiKASA-3DVG  & \textit{3D-VG} & $25.7  $ & $3.6  $ & $14.2  $ & $0.0  $ & $18.8  $ & $0.0  $ & $25.1  $ & $6.1  $ & $18.7  $ & $6.9  $ & $21.6  $ & $5.3  $ \\
ViewRefer  & \textit{3D-VG} & $27.6  $ & $4.5  $ & $22.7  $ & $2.9  $ & $22.3  $ & $0.0  $ & $31.5  $ & $9.9  $ & $24.2  $ & $11.7  $ & $26.7  $ & $8.8  $ \\
% Vil3DRef  & \textit{3D-VL} & $29.3  $ & $5.6  $ & $20.7  $ & $0.0  $ & $20.9  $ & $0.0  $ & $31.2  $ & $7.9  $ & $23.0  $ & $9.6  $ & $26.3  $ & $7.4  $ \\
3D-VisTA  & \textit{3D-VG} & $26.9  $ & $4.7  $ & $23.7  $ & $2.2  $ & $22.8  $ & $4.7  $ & $30.8  $ & $9.0  $ & $25.3  $ & $10.3 $ & $26.9  $ & $8.0  $ \\
PQ3D      & \textit{3D-VG} & $29.7  $ & $4.1  $ & $24.6  $ & $2.9  $ & $23.2  $ & $0.0  $ & $34.6  $ & $8.6  $ & $24.4  $ & $9.7  $ & $28.2  $ & $7.5  $ \\  
GPT-4o  & \textit{LLM} & $44.6  $& $13.5  $ & $31.3  $ & $5.8  $ & $28.1  $ & $7.0  $ & $30.7  $ & $12.0  $ & $21.4  $ & $8.8  $ & $30.0  $ & $10.3 $ \\
% Qwen2 w/ pred labels  & \textit{LLM} & $42.5  $ & $10.2  $ & $29.3  $ & $6.5  $ & $25.0  $ & $4.6  $ & $28.7  $ & $10.6  $ & $20.9  $ & $6.7  $ & $28.5  $ & $8.3  $ \\
Qwen2.5  & \textit{LLM} & $47.0  $ & $14.6  $ & $34.2  $ & $5.1  $ & $26.8  $ & $4.7  $ & $34.8  $ & $15.1  $ & $23.9  $ & $9.5  $ & $32.8  $ & $11.5  $ \\
DeepSeek V3  & \textit{LLM} & $48.5  $ & $14.9  $ & $35.1  $ & $6.5  $ & $31.7  $ & $4.7  $ & $34.0  $ & $15.1  $ & $24.3  $ & $9.6  $ & $33.3  $ & $11.7  $ \\
DeepSeek R1  & \textit{LLM} & $54.9  $ & $17.4  $ & $39.8  $ & $8.0  $ & $33.0  $ & $7.0  $ & $42.1  $ & $18.3  $ & $31.4  $ & $11.8  $ & $40.2  $ & $14.2$ \\
% \rowcolor[gray]{0.7} GPT-4o w/ GT labels  & \textit{LLM} & $69.2  $ & $38.1  $ & $73.1  $ & $37.0  $ & $73.7  $ & $32.6  $ & $73.2  $ & $48.3  $ & $75.9  $ & $52.1  $ & $73.4  $ & $46.6  $ \\
Chat-Scene  & \textit{3D LLM} & $43.3  $ & $10.2  $ & - & - & - & - & - & - & - & - & - & - \\
GPT4Scene  & \textit{LVLM} & $56.0  $ & $18.7  $ & - & - & - & - & - & - & - & - & - & - \\
% GPT4Scene  & \textit{LVLM} & $51.2  $ & $14.0  $ & - & - & - & - & - & - & - & - & - & - \\
\midrule
\rowcolor[gray]{0.9} \textbf{Fine-tune} & & & & & &  & & & & & & & \\
MiKASA-3DVG  & \textit{3D-VG} & 57.8          & 20.3          & 53.0 & 10.9          & 48.7          & 2.3                     & 66.4          & 33.6          & 57.2          & 30.6          & 59.1          & 26.9          \\
ViewRefer  & \textit{3D-VG} & 59.9          & 20.8          & 54.6          & 6.5           & 48.7          & 4.7                     & 68.2          & 34.8          & 57.3          & 30.0          & 60.2          & 26.8          \\
% Vil3DRef  & \textit{3D-VL} & 60.2          & 20.8          & 53.3          & 11.6          & 53.6 & \textbf{11.6}    & \textbf{70.1} & 37.5          & 58.0          & 29.7          & 61.1          & 27.8          \\
3D-VisTA  & \textit{3D-VG} & $57.6 $ & $21.0 $ & $54.3 $ & $15.9 $ & $49.7 $ & $7.8 $ & $69.7 $ & $38.4 $ & $59.3 $ & $32.4 $ & $60.9 $ & $29.5 $ \\
% 3D-VisTA  & \textit{3D-VL} & $57.6 \pm 0.5$ & $21.0 \pm 0.2$ & $54.3 \pm 0.6$ & $15.9 \pm 4.0$ & $49.7 \pm 1.3$ & $7.8 \pm 3.5$ & $69.7 \pm 0.3$ & $38.4 \pm 0.2$ & $59.3 \pm 0.5$ & $32.4 \pm 0.8$ & $60.9 \pm 0.3$ & $29.5 \pm 0.2$ \\
PQ3D & \textit{3D-VG} & $ 54.8  $ & $17.8  $ & $49.3  $ & $9.9  $ & $46.4  $ & $4.7  $ & $ 65.2  $ & $32.1  $ & $56.1 $ & $30.0  $ & $57.3  $ & $25.7  $ \\
% PQ3D & \textit{3D-VL} & $ 54.8  \pm 0.8$ & $17.8  \pm 0.7$ & $49.3  \pm 1.3$ & $9.9  \pm 2.5$ & $46.4  \pm 2.1$ & $4.7  \pm 0.0$ & $ 65.2  \pm 0.5$ & $32.1  \pm 0.7$ & $56.1 \pm 0.3$ & $30.0  \pm 0.7$ & $57.3  \pm 0.1$ & $25.7  \pm 0.4$ \\
% Chat-Scene & \textit{3D LLM} & $ 45.5  $ & $ 13.1 $ & - & - & - & - & - & - & - & - & - & - \\
\midrule
\textbf{\model} (ours) & \textit{3D LLM} & $ \mathbf{61.6}  $ & $\mathbf{26.3}  $ & $\mathbf{57.1}  $ & $\mathbf{18.5}  $ & $\mathbf{54.8}  $ & $\mathbf{10.4}  $  & $\mathbf{68.9}  $ & $\mathbf{42.3}  $ & $\mathbf{61.9}  $ & $\mathbf{37.1}  $ & $\mathbf{63.1}  $ & $\mathbf{33.9}  $  \\ 
% \model & \textit{3D LLM} & $ \mathbf{61.6}  \pm 0.2$ & $\mathbf{26.3}  \pm 1.7$ & $\mathbf{57.1}  \pm 0.8$ & $\mathbf{18.5}  \pm 2.4$ & $\mathbf{54.8}  \pm 0.4$ & $\mathbf{10.4}  \pm 2.1$  & $ 68.9  \pm 1.2$ & $\mathbf{42.3}  \pm 1.2$ & $\mathbf{61.9}  \pm 0.3$ & $\mathbf{37.1}  \pm 0.7$ & $\mathbf{63.1}  \pm 0.2$ & $\mathbf{33.9}  \pm 0.5$  \\  
\bottomrule
\end{tabular}
}
\end{table*}

\begin{table}[t]
% \vspace{-5mm}
\centering
 \caption{Results on \datasetnav. ``s-SR'' denotes the average success rate over all steps. ``t-SR'' denotes the average success rate over all tasks. A task is considered successful if all steps of the task episode succeed. 
 % We ran each experiment three times and reported error bars.
 }
    \label{tab:nav_result}
\begin{tabular}{lccc}
\toprule
 & \textbf{s-SR (\%)} & \textbf{t-SR (\%)} & \textbf{SPL (\%)} \\
\midrule
\rowcolor[gray]{0.9} \textbf{Zero-shot} & & & \\
End-to-end & 5.2 & 1.2 & 4.8 \\
Modular & 14.7 & 3.8 & 10.2 \\
% Modular w/o context & 14.0 & 1.5 & 8.2 \\
\midrule
\rowcolor[gray]{0.9} \textbf{Fine-tune} & & & \\
End-to-end & $12.1 $ & $7.7 $ & $10.1 $ \\
% End-to-end (Fine-tune) & $12.1 \pm 1.7$ & $7.7 \pm 0.7$ & $10.1 \pm 1.2$ \\
\bottomrule
\end{tabular}
\end{table}

\begin{figure}[htbp]
    \centering
    \includegraphics[width=\columnwidth]{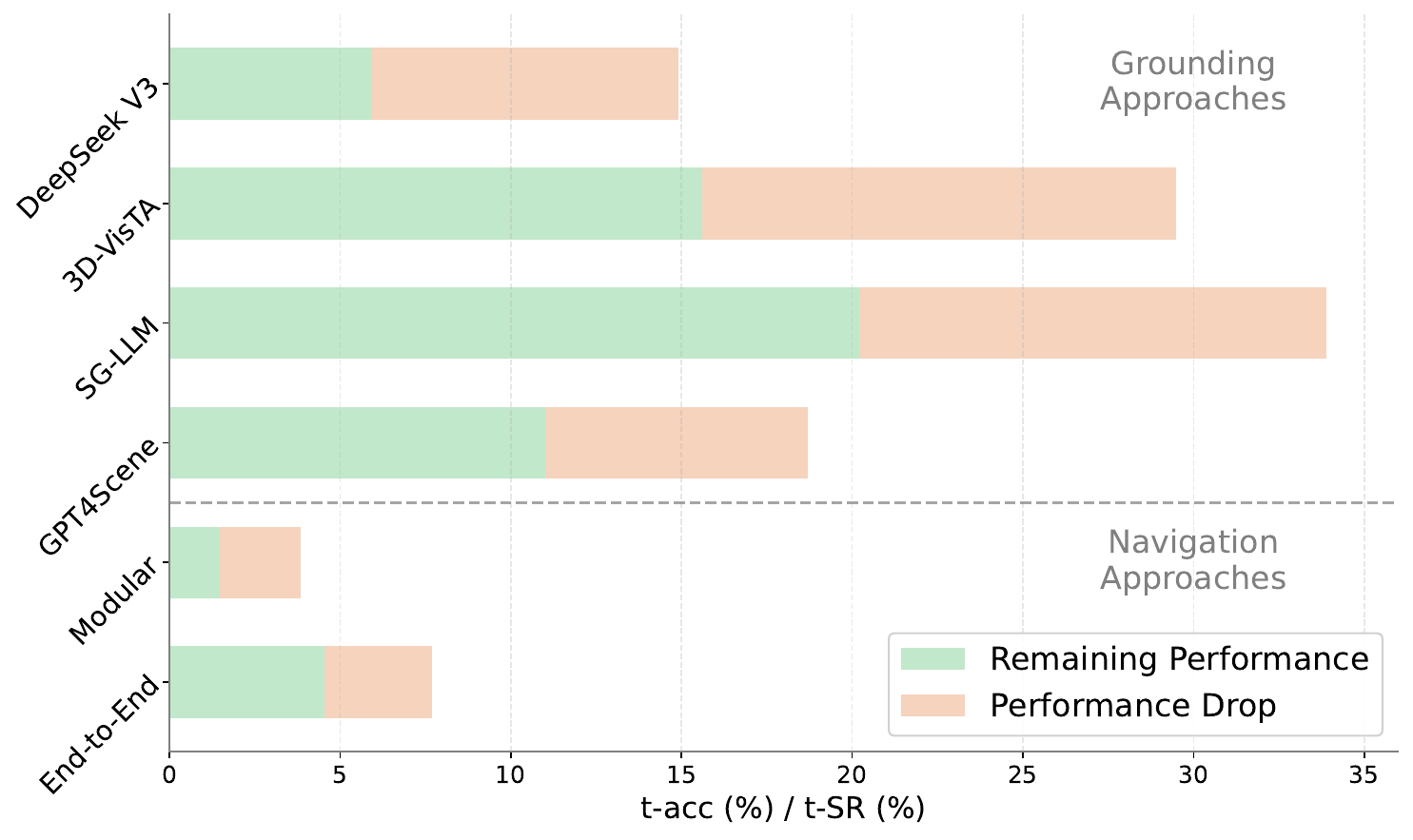}
    \caption{The stacked bar chart shows the performance degradation of grounding and navigation methods when sequential context $t, \{s_1, \dots, s_{i-1}\}$ is removed. For each bar, green segment shows the retained performance without context, while red segment shows the performance drop caused by context removal.}
    % The left Y-axis (blue bars) shows the absolute performance drop in percentage points (pp), while the right Y-axis (orange bars) represents the relative performance drop (\%). A vertical dashed line separates the grounding and the navigation approaches.
    \label{fig:ablate-seq}
\end{figure}

% \begin{figure}[!h]
%     \centering
%     \includegraphics[width=\linewidth]{figures/data-amount.pdf}
%     \caption{Impact of training data volume and data efficiency comparison.}
%     \label{fig:ablate-data}
% \end{figure}

% \begin{figure*}[!h]
% \begin{minipage}[5]{0.5\linewidth}
%     \centering
%     % \includegraphics[width=\textwidth]{figures/sequential-ablation.pdf}
%     \includegraphics[width=\textwidth]{figures/context_ablation_figure.pdf}
%     \caption{Ablation of contextual information.}
%     \label{fig:ablate-seq}
% \end{minipage}
% \hfill
% \begin{minipage}[5]{0.5\linewidth}
%     \centering
%     \includegraphics[width=\textwidth]{figures/data-amount.pdf}
%     \caption{Impact of training data volume and data efficiency comparison.}
%     \label{fig:ablate-data}
% \end{minipage}
% \end{figure*}

\subsubsection{Effect of Removing Sequential Context}

We investigate the impact of removing sequential context $t, \{s_1, \dots, s_{i-1}\}$ during the processing of step $s_i$. Results in \cref{fig:ablate-seq} show significant performance drops in task accuracy/task success rate for both grounding and navigation approaches. For grounding approaches, task accuracy drops by 7.7--13.9\% (40.4\%--60.4\% relative). Navigation approaches exhibit smaller absolute drops (2.4--3.2\%) but critical relative declines of >40\%. These findings indicate that the sequential context is crucial for both grounding and navigation tasks while highlighting variations in how effectively different approaches leverage it.

% \paragraph{Effect of offering contextual information.} To investigate the impact of contextual information, we eliminate multi-step action context during both the training and inference phases, providing only the task description and current action step. The experimental results illustrated in \cref{fig:ablate-seq} reveal a significant decline in task accuracy upon the removal of contextual information for both LEO and 3D-VisTA. Specifically, LEO exhibits an average t-acc drop of 3.4\%, while 3D-VisTA demonstrates an even more pronounced decline of 5.0\%. This suggests that the models have, to a certain extent, learned to leverage contextual information during the grounding process. In contrast, PQ3D experiences a more modest performance reduction, with an average t-acc decrease of only 0.8\%. This limited decline can be attributed to PQ3D's reliance on a CLIP text encoder, which struggles to interpret lengthy sentences~\citep{zhang2024long}, thereby leading to overfitting on shorter, single-step instructions.

% \paragraph{Impact of training data volume and data efficiency comparison.} \cref{fig:ablate-data} shows that increasing the volume of training data utilized during fine-tuning improves the performance of all three models. Notably, LEO exhibits superior data efficiency, achieving comparable performance to PQ3D and 3D-VisTA using only 25\% of the data. This advantage is likely attributable to LEO's foundation on a large language model, which has been pre-trained on a vast array of task-relevant information and acquired common-sense knowledge.

\subsection{Qualitative Results}
% \begin{figure*}[]
%     \centering
%     \begin{subfigure}{0.45\linewidth}
%         \includegraphics[width=\textwidth]{figures/qualitative_nav_upper.pdf}
%     \end{subfigure}
%     % \vspace{0.5cm}
%     \begin{subfigure}{0.45\linewidth}
%         \includegraphics[width=\textwidth]{figures/qualitative_nav_lower.pdf}
%     \end{subfigure}
%     \caption{Qualitative result from modular navigation model.}
%     \label{fig:qualitative-nav}
% \end{figure*}

\begin{figure*}[htbp]
    \centering
    \includegraphics[width=\textwidth]{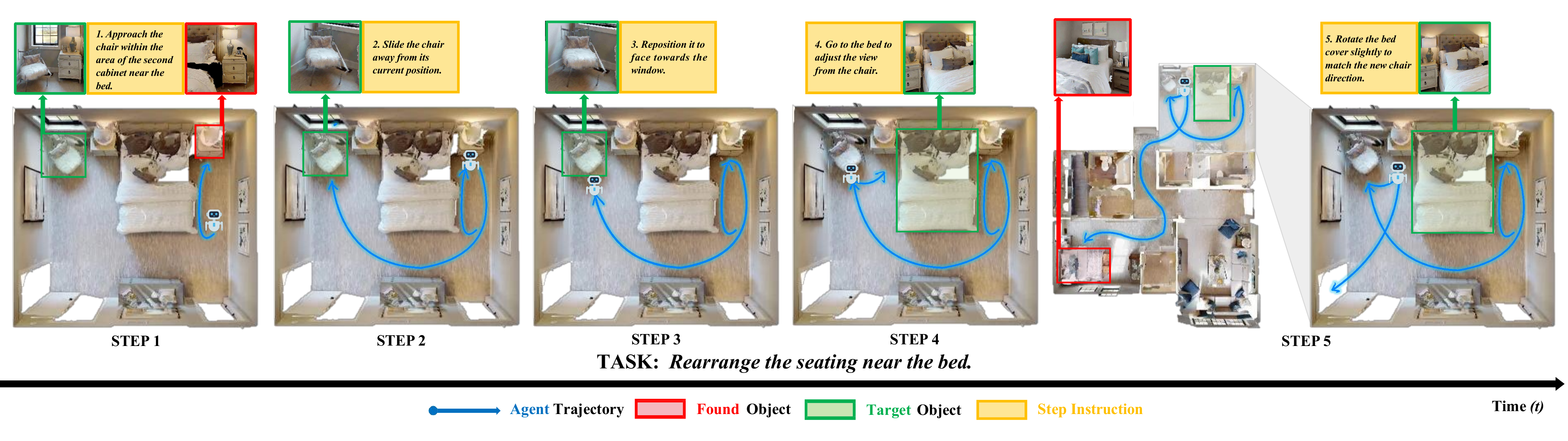}
    \caption{Qualitative result from modular navigation agent.}
    \label{fig:qualitative-nav}
\end{figure*}

\cref{fig:qualitative-nav} shows a task episode from the \datasetnav benchmark, highlighting challenges and complexities in task-oriented sequential grounding and navigation. In Step 1, the modular agent erroneously attends to a cabinet instead of the specified target (``chair''), suggesting susceptibility to semantic confusion during target interpretation from task-oriented steps. In Step 5, the agent fails to maintain consistency and selects an alternative bed observed during exploration.

%% file: sec/6_conclusion.tex
\section{Conclusion}
\label{sec:conclusion}
This work introduces the task of Task-oriented Sequential Grounding and Navigation in 3D Scenes, along with \dataset, a large-scale dataset designed to advance research in this domain. Through extensive evaluations of grounding and navigation approaches on \dataset, we underscore the significant challenges posed by sequential grounding and navigation tasks, particularly in understanding the sequential context in task-oriented steps. Additionally, we propose a state-of-the-art model, \model, for the sequential grounding task, leveraging a stepwise grounding mechanism to address these challenges. We encourage the research community to shift focus from traditional grounding and navigation studies to task-oriented scenarios, fostering the development of more sophisticated, context-aware embodied agents.

% Our findings underscore the need for further research to develop models capable of effectively interpreting task-oriented instructions and leveraging sequential context. We encourage the research community to shift focus from traditional grounding and navigation studies towards more practical, task-oriented applications, ultimately driving the development of more sophisticated and capable embodied agents.

% These results emphasize the necessity for further research and model development. We encourage the community to move beyond traditional 3D visual grounding towards more practical, task-oriented applications, paving the way for more advanced and capable embodied agents.

%% file: appendix.tex
% \newpage
\clearpage
\appendix

\renewcommand\thefigure{A\arabic{figure}}
\setcounter{figure}{0}
\renewcommand\thetable{A\arabic{table}}
\setcounter{table}{0}
\renewcommand\theequation{A\arabic{equation}}
\setcounter{equation}{0}
\pagenumbering{arabic}% resets `page` counter to 1
\renewcommand*{\thepage}{A\arabic{page}}
\setcounter{footnote}{0}

\section{Appendix}
\subsection{Details of Dataset Construction}
\label{sec:dataset construction details}

\textbf{Detailed prompt design for task generation.}
The prompts used in the task generation process are illustrated in \cref{fig:prompt messages}, with the ``System prompt'' specifically detailed in \cref{fig:system prompt}. Within the system prompt, in-context examples, labeled as ``<EXAMPLES>'', are provided in \cref{fig:examples}. Notably, we deliberately omit to show GPT-4 the corresponding scene graph for the provided response examples, as an overly long context increases the likelihood of errors.

\begin{figure*}[th!]
\centering
\begin{minipage}{0.9\linewidth}\vspace{0mm}    \centering
\begin{tcolorbox}
\fontsize{7.0pt}{0.8\baselineskip}\selectfont
messages = [\{`role': `system', `content': \hyperref[fig:system prompt]{\blueprompt{System prompt}}\}, \{`role': `user', `content': \blueprompt{Scene graph of the scene to process}\}]
\end{tcolorbox}
\caption{Prompts messages for GPT-4 task generation.}
\label{fig:prompt messages}
\end{minipage}
\end{figure*}

\noindent\textbf{Ensuring contextual richness in generated tasks.}
To enhance the contextual richness of generated tasks, we explicitly instruct GPT-4 to incorporate pronouns such as ``it,'' ``them,'' ``here,'' and ``the other'' (\cref{fig:system prompt}). We also curate in-context examples (\cref{fig:examples}) to illustrate typical cases. For instance, in the task ``Make me a cup of coffee,'' step 5, ``Put the cup down,'' and step 8, ``Go back to the table,'' both refer to the table introduced in step 4, ``Walk to the table close to a cabinet.'' Similarly, in the task ``Watch TV from the sofa,'' step 2, ``Grab the black remote lying on it,'' references the table mentioned in step 1, ``Go to the black table to the left of the fire extinguisher.'' Later, step 5, ``Place the remote here,'' refers to the table described in step 4, ``Walk to the table in the middle of the bed and the white cabinet,'' while step 7, ``Sit here to admire the TV show,'' points to the sofa reached in step 6, ``Walk to the black sofa close to the bed.'' In the task ``Browse the internet,'' step 3, ``Sit down on the nearest chair,'' references the chair near the computer tower mentioned in step 2, ``Turn on the computer tower behind the desk and the bookshelf.'' These strategies ensure that the generated tasks are contextually rich, with a natural flow and strong continuity.

\noindent\textbf{Post-generation process.} After generating the tasks, we filter out any outputs with formatting errors and meticulously verify that all assigned targets are exactly present in the corresponding scenes. Moreover, we observe that tasks exceeding ten steps often introduce hallucinated objects or problematic step instructions, which can negatively impact dataset quality. As a result, we discard any tasks containing more than ten steps.

\begin{figure*}[th!]
\centering
\begin{minipage}{0.9\linewidth}\vspace{0mm}    \centering
\begin{tcolorbox}
\fontsize{7.0pt}{0.8\baselineskip}\selectfont
You are a helpful assistant that can generate diverse tasks in an indoor scene.\\

The scene is represented by a scene graph in the JSON dictionary format. Each entity in the scene graph denotes an object instance, named `<category>-<ID>'. The `caption' describes the object's attributes, such as `color', `material', etc. The `relations' describes the object's spatial relations with other objects. For example, from the scene graph:\\
\textasciigrave\textasciigrave\textasciigrave\\
{`sofa-1': {`relations': [`to the right of armchair-2', `in front of table-3'], `caption': `Grey velvet sofa with a rectangular shape and a back and arms, suitable for use in a living room.'}, `armchair-2': {`relations': [`to the left of sofa-1'], `caption': `The armchair is made of leather, specifically black leather, and has a spherical shape.'}, `table-3': {`relations': [], `caption': `The table is a rectangular wooden table with a brown finish, sometimes used as a dining table or coffee table, with a smooth wooden texture and various styles, including a sign or place setting on it, and can have plates or a white cloth on it.'}}\\
\textasciigrave\textasciigrave\textasciigrave\\

You can know that `sofa-1' is grey, the `armchair-2' is made of leather, the `table-3' is made of wood, the `armchair-2' is on the left of the `sofa-1', the `sofa-1' is in front of the `table-3'.\\

Using the provided scene graph, design daily tasks that a person can do in this scene. Besides, decomposing every task into a sequence of steps that can be performed using the objects in this scene. For each step, give the target object that the person should attend to. Your output must follow the template below:\\
\textasciigrave\textasciigrave\textasciigrave\\
Task: \#Describe the task using one sentence.\#\\
Steps:\\
1. \#The step must perform only one action. Split actions such as `pick up xxx and place it xxx' into two separate steps. All objects, attributes, and relations must be explicitly listed in the given scene graph. Do not include the IDs of the objects, use ordinal words, attributes, and relations to refer to different object instances of the same category. Use pronouns (`it', `them', `here', and `the other', etc.) as much as possible to make the step concise.\# [\#Use `<category>-<ID>' to denote the target object. Do NOT assume objects that do not exist in the scene graph! Each step must have exactly one target object. \#]\\
2. ...\\
3. ...\\
...\\
\textasciigrave\textasciigrave\textasciigrave\\

Here are some examples:\\
\textasciigrave\textasciigrave\textasciigrave\\
\hyperref[fig:examples]{\blueprompt{<EXAMPLES>}}\\
\textasciigrave\textasciigrave\textasciigrave\\

Generate 5 different tasks involving different objects and separate these tasks by "===".
\end{tcolorbox}
\caption{\blueprompt{System prompt} for GPT-4 task generation.}
\label{fig:system prompt}
\end{minipage}
\end{figure*}

\begin{figure*}[th!]
\centering
\begin{minipage}{0.9\linewidth}\vspace{0mm}    \centering
\begin{tcolorbox}
\fontsize{7.0pt}{0.8\baselineskip}\selectfont
Task: Make me a cup of coffee.\\
Steps:\\
1. Go to the long desk against the wall. [desk-15]\\
2. Choose a cup from those white, plastic cups on the top of the desk. [cups-19]\\
3. Fill it with coffee at the coffee maker. [coffee maker-16]\\
4. Walk to the table close to a cabinet. [table-23]\\
5. Put the cup down. [table-23]\\
6. Return to the long desk. [desk-15]\\
7. Fetch a plate from a bunch of steel plates below a picture frame hanging on the wall. [plates-17]\\
8. Go back to the table. [table-23]\\
9. Put the cup on the plate on the table. [table-23]\\
===\\
Task: Watch tv from the sofa.\\
Steps:\\
1. Go to the black table to the left of the fire extinguisher. [table-30]\\
2. Grab the black remote lying on it. [remote-36]\\
3. Turn on the tv with the remote. [tv-38]\\
4. Walk to the table in the middle of the bed and the white cabinet. [table-58]\\
5. Place the remote here. [table-58]\\
6. Walk to the black sofa close to the bed. [sofa-14]\\
7. Sit here to admire tv show. [sofa-14]\\
===\\
Task: Clean the mirror.\\
Steps:\\
1. Walk to the white cabinet. [cabinet-7]\\
2. Grab the towel on it. [towel-10]\\
3. Put the towel into the sink. [sink-37]\\
4. Turn the faucet on. [faucet-13]\\
5. Wet the towel in the sink. [sink-37]\\
6. Turn the faucet off. [faucet-13]\\
7. Wipe the mirror with the towel. [mirror-11]\\
8. Put the towel into the sink again. [sink-37]\\
9. Turn the faucet on. [faucet-13]\\
10. Wash the towel in the sink. [sink-37]\\
11. Turn the faucet off. [faucet-13]\\
12. Wring the towel dry in the sink. [sink-37]\\
13. Put it back to the cabinet. [cabinet-7]\\
===\\
Task: Browse the internet.\\
Steps:\\
1. Walk to the desk adorned with papers. [desk-19]\\
2. Turn on the computer tower behind the desk and the bookshelf. [computer tower-7]\\
3. Sit down on the nearest chair. [chair-26]\\
4. Fetch the mouse on the desk. [mouse-8]\\
5. Look at the screen of the monitor. [monitor-14]\\
===\\
Task: Go to sleep.\\
Steps:\\
1. Go to the curtain. [curtain-11]\\
2. Close it. [curtain-11]\\
3. Walk to the nightstand with the telephone. [nightstand-15]\\
4. Turn off the lamp on this nightstand. [lamp-19]\\
5. Go to the other nightstand. [nightstand-14]\\
6. Set the alarm on it. [alarm clock-28]\\
7. Lie down on the bed. [bed-20]
\end{tcolorbox}
\caption{\blueprompt{<EXAMPLES>} in system prompt for GPT-4 task generation.}
\label{fig:examples}
\end{minipage}
\end{figure*}

\noindent\textbf{Details in navigation episode construction.} We construct the navigation dataset \datasetnav following the GOAT-Bench~\citep{khanna2024goat} setting. Each task episode's starting position is randomly sampled such that the distance to the target object for the first step $s_1$ lies between 1--30 meters. We filter episodes that involve target objects exhibiting an IoU (Intersection over Union) score < 0.05 across all viewpoints within a 1-meter radius of its location.

\begin{figure*}[!t]
    \centering
    \includegraphics[width=\linewidth]{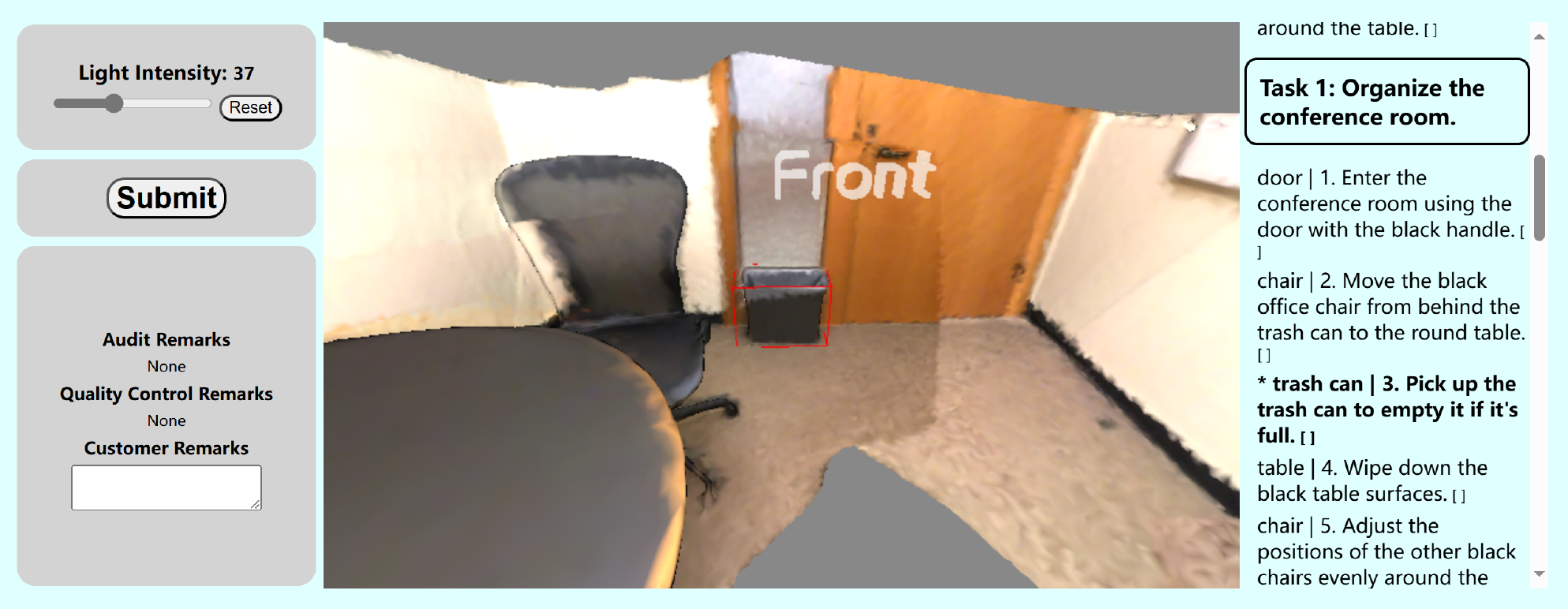}
    \caption{Screenshot of the interface for human verification.}
    \label{fig:interface}
\end{figure*}

\noindent\textbf{Details in human verification}
\cref{fig:interface} shows the interface used for human verification. The interface consists primarily of an interactive 3D mesh window and a right-hand column that displays task data. When a specific step is selected, the target object is highlighted within the 3D mesh using a red bounding box. Users can rotate, translate, and zoom in or out within the 3D mesh window to inspect the scene thoroughly. Annotators evaluate each step by marking it with a tick (correct) or a cross (incorrect). Following this verification process, tasks containing one incorrect step are manually revised, while those with multiple incorrect steps are discarded to maintain the overall quality of the dataset.

\subsection{Additional Data Statistics and Analysis}
\textbf{Training and evaluation splits.} \dataset is devided into training and evaluation sets. For 3RScan, scenes from its training and evaluation splits are used to construct our training set, while scenes from its test split were designated for evaluation. For the other datasets, we adhere to the original split of the 3D scenes as provided. The task and step count statistics for the training and evaluation splits are detailed in \cref{tab:dataset_stats}.

\begin{table}[ht]
\centering
\caption{Statistics of the training and evaluation splits for various datasets in \dataset.}
\label{tab:dataset_stats}
\resizebox{1\columnwidth}{!}{
\begin{tabular}{lcccc}
\toprule
& & Training Set & Evaluation Set & Train+Eval \\ 
\midrule
\multirow{2}{*}{3RScan} & \# tasks & 2,056 & 138 & 2,194 \\ 
& \# steps & 10,622 & 696 & 11,318 \\ 
\midrule
\multirow{2}{*}{ScanNet} & \# tasks & 2,731 & 443 & 3,174 \\ 
& \# steps & 13,634 & 2,108 & 15,742 \\ 
\midrule
\multirow{2}{*}{MultiScan} & \# tasks & 504 & 43 & 547 \\ 
& \# steps & 2,459 & 224 & 2,683 \\ 
\midrule
\multirow{2}{*}{ARKitScenes} & \# tasks & 6,952 & 443 & 7,395 \\ 
& \# steps & 37,552 & 2,335 & 39,887 \\ 
\midrule
\multirow{2}{*}{HM3D} & \# tasks & 8,146 & 890 & 9,036 \\ 
& \# steps & 38,833 & 3,873 & 42,706 \\ 
\midrule
\multirow{2}{*}{Total} & \# tasks & 20,389 & 1,957 & 22,346 \\ 
& \# steps & 103,100 & 9,236 & 112,336 \\ 
\bottomrule
\end{tabular}
}
\end{table}

\begin{figure*}
  \centering
  % \vspace{-10mm}
  \begin{subfigure}{0.32\linewidth}
    % \fbox{\rule{0pt}{2in} \rule{.9\linewidth}{0pt}}
    \includegraphics[width=\linewidth]{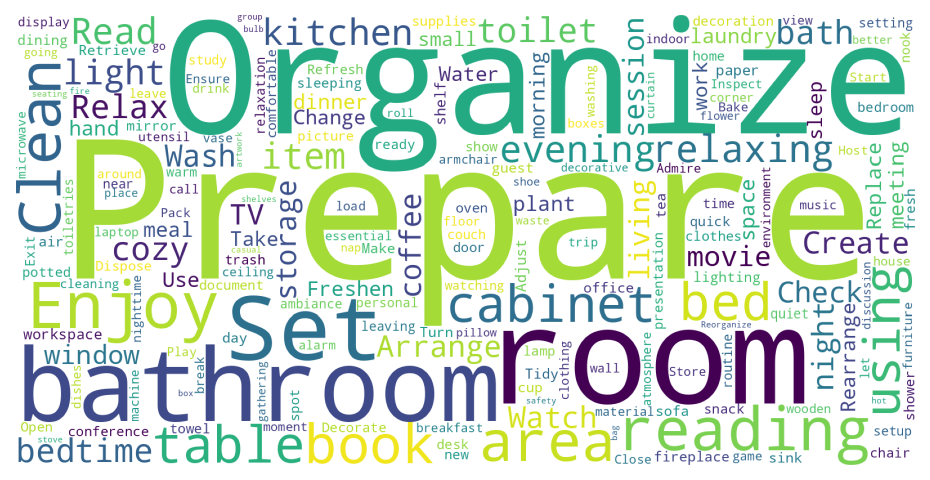}
    \caption{Task Description.}
    \label{fig:task_descriptions_wordcloud}
  \end{subfigure}
  \hfill
  \begin{subfigure}{0.32\linewidth}
    % \fbox{\rule{0pt}{2in} \rule{.9\linewidth}{0pt}}
    \includegraphics[width=\linewidth]{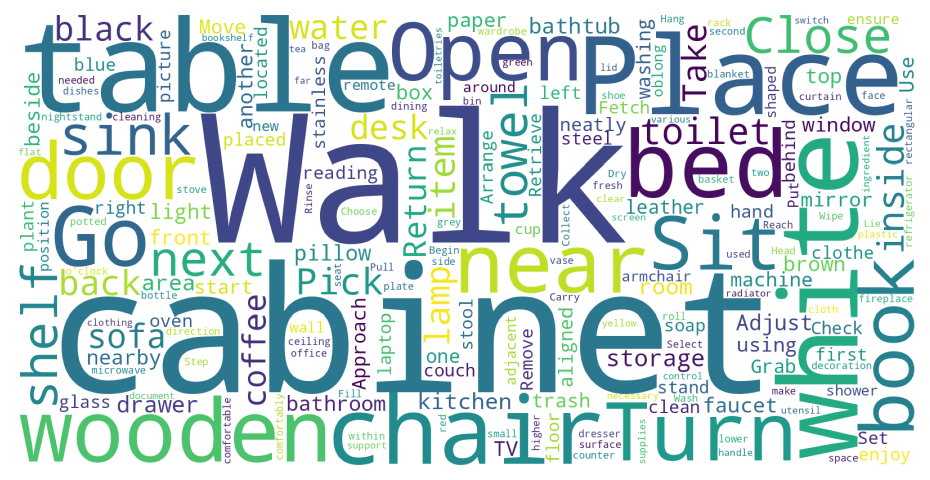}
    \caption{Step-by-step Instructions.}
    \label{fig:action_steps_wordcloud}
  \end{subfigure}
  \hfill
  \begin{subfigure}{0.32\linewidth}
    % \fbox{\rule{0pt}{2in} \rule{.9\linewidth}{0pt}}
    \includegraphics[width=\linewidth]{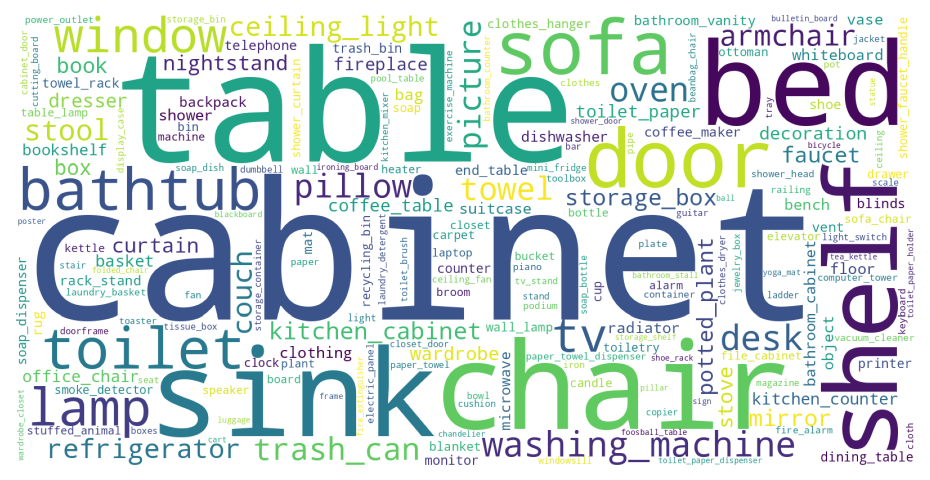}
    \caption{Target Object Categories.}
    \label{fig:labels_wordcloud}
  \end{subfigure}
  \caption{Word clouds of (a) task description, (b) step-by-step instructions, and (c) target object categories.}
  \label{fig:word_clouds}
\end{figure*}

\noindent\textbf{Word cloud analysis.} To visually represent the linguistic diversity of \dataset, we generate three word clouds (see \cref{fig:word_clouds}). \Cref{fig:task_descriptions_wordcloud} and \cref{fig:action_steps_wordcloud} illustrate word frequency distributions in task descriptions and step instructions, respectively.  Analysis reveals that task descriptions prioritize abstract, goal-oriented verbs, with ``prepare'' and ``organize'' emerging as dominant terms. In contrast, step instructions feature granular, executable directives: ``walk'' and ``place'' rank as predominant action verbs, while ``table'' and ``white'' appear as the most frequently referenced object and adjective, respectively. This contrast highlights how task descriptions frame high-level objectives, whereas step instructions decompose them into contextually grounded operations. Finally, \cref{fig:labels_wordcloud} identifies recurring target object categories in the dataset, encompassing ``cabinet,'' ``table,'' ``chair,'' ``sink,'' ``bed,'' and ``shelf,'' reflecting the diversity of household objects commonly involved in procedural task guidance.

\noindent\textbf{Quantifying context-dependency in \dataset.} To evaluate the prevalence of context-dependent reasoning, we performed a manual annotation study on 100 randomly sampled tasks from the evaluation set. Expert annotators were provided with task descriptions, step-by-step instructions, and interactive 3D scenes rendered in a web interface. They were instructed to flag steps requiring contextual disambiguation—defined as situations where the target object could not be uniquely identified from candidates with just the current step, necessitating inference from the task description or prior step history. Analysis revealed that 20.7\% of all steps involved context-dependent disambiguation, with 57.0\% of tasks containing at least one such step.

\subsection{Details of Grounding Baselines}

\noindent\textbf{3D-VG Models.} 3D-VG models take point clouds as the basic scene representation for visual grounding tasks. For a fair comparison, we use point clouds as the scene representation and employ the same PointNet++~\citep{pointnet++} encoder to extract scene features across all 3D-VG baselines. Among 3D-VG baselines, MiKASA-3DVG~\citep{chang2024mikasa} and ViewRefer~\citep{guo2023viewrefer} are specialized models for 3D visual grounding, while 3D-VisTA~\citep{3d-vista} and PQ3D~\citep{pq3d} are unified 3D vision-language models, which are capable of handling diverse 3D vision-language tasks, such as 3D grounding, question answering, and captioning.
% We train all 3D-VL baseline models for 50 epochs across all available datasets in \datasets.

\noindent\textbf{LLMs.} LLM methods leverage PointNet++~\citep{pointnet++}, pre-trained on ScanNet, as the classifier to predict a semantic category for each object instance in the scene based on their ground-truth masks. The LLM receives structured scene information in JSON format, including each object's ID, predicted semantic category, and bounding box coordinates calculated from ground-truth masks. Given the scene information JSON, task description $t$, and step-by-step instructions $s_1,...s_n$, the LLM generates a list of target object IDs. \cref{table::prompt_for_gpt_baseline} details the prompt messages employed in the LLM methods.

\noindent\textbf{3D LLM.} Chat-Scene enhances 3D LLMs by incorporating visual grounding capabilities through unique ID tokens assigned to each object instance. These ID tokens are associated with vision feature tokens through a next-token prediction framework.
% To ensure a fair comparison with 3D-VG models, we employ point clouds as the scene representation for Chat-Scene.
Experiments are conducted on the ScanNet split of \dataset, as the official implementation does not support preprocessing for other 3D scene datasets.

\noindent\textbf{Large Vision-Language Model.} GPT4Scene streamlines the 3D visual grounding process by leveraging only 3D positional information to annotate video frames and a bird’s-eye-view (BEV) image with object instance IDs. These annotated inputs are then processed through a large vision-language model backbone to generate responses for the 3D visual grounding task. Similar to Chat-Scene, experiments for GPT4Scene are performed on the ScanNet split of \dataset, as the preprocessing code for video frames in other 3D scene datasets is currently unavailable.

\begin{figure*}[th!]
\centering
\begin{minipage}{0.9\linewidth}\vspace{0mm}    \centering
\begin{tcolorbox}
\fontsize{7.0pt}{0.8\baselineskip}\selectfont
\# system prompt (role: system)\\
You are tasked with identifying the target object for each step in a given task. Each scene contains various objects, and your response should provide the target object for each step in the format <label-id>, maintaining the sequence of steps. For example:\\
\\
\# example task (role: user)\\
Task: Make me a cup of coffee and serve it on a plate.\\
Steps:\\
1. Go to the long desk against the wall.\\
2. Fetch a plate from a bunch of steel plates below the picture frame.\\
3. Walk to the table close to a cabinet.\\
4. Put the plate on it.\\
5. Return to the long desk.\\
6. Choose a cup from those white, plastic cups on the desk.\\
7. Fill it with coffee at the coffee maker.\\
8. Go back to the table.\\
9. Put down the cup of coffee.\\
\\
\# example scene (role: user)\\
{\\
    "table-24": {\\
        "position": [\\
            -4.91,\\
            2.25,\\
            -0.97\\
        ],\\
        "size": [\\
            2.03,\\
            1.25,\\
            0.84\\
        ]\\
    },\\
    ...\\
}\\
\\
\# example response (role: assistant)\
1. desk-15\\
2. plates-17\\
3. table-23\\
4. table-23\\
5. desk-15\\
6. cups-19\\
7. coffee maker-16\\
8. table-23\\
9. table-23\\
\\
\# role: user\\
< CURRENT TASK \& SCENE >
\end{tcolorbox}
\caption{Prompt messages used in LLM methods.}
\label{table::prompt_for_gpt_baseline}
\end{minipage}
\end{figure*}

\subsection{Training Details}
All 3D-VG models and \model are trained across all available datasets in \dataset for 50 epochs. We employ the AdamW optimizer with a learning rate of 1e-4, \(\beta_1=0.9\), \(\beta_2=0.999\), and a weight decay of 0.05. For 3D-VG models, we use a batch size of 32; for \model, the batch size is reduced to 16 due to GPU memory constraints. We use LoRA tuning~\citep{hu2022lora} for the parameters of the LLM in \model with a rank of 16. We train the end-to-end navigation policy for 70M steps until convergence.

\subsection{Human Study}
\begin{figure*}[htbp]
    \centering
        \includegraphics[width=\linewidth]{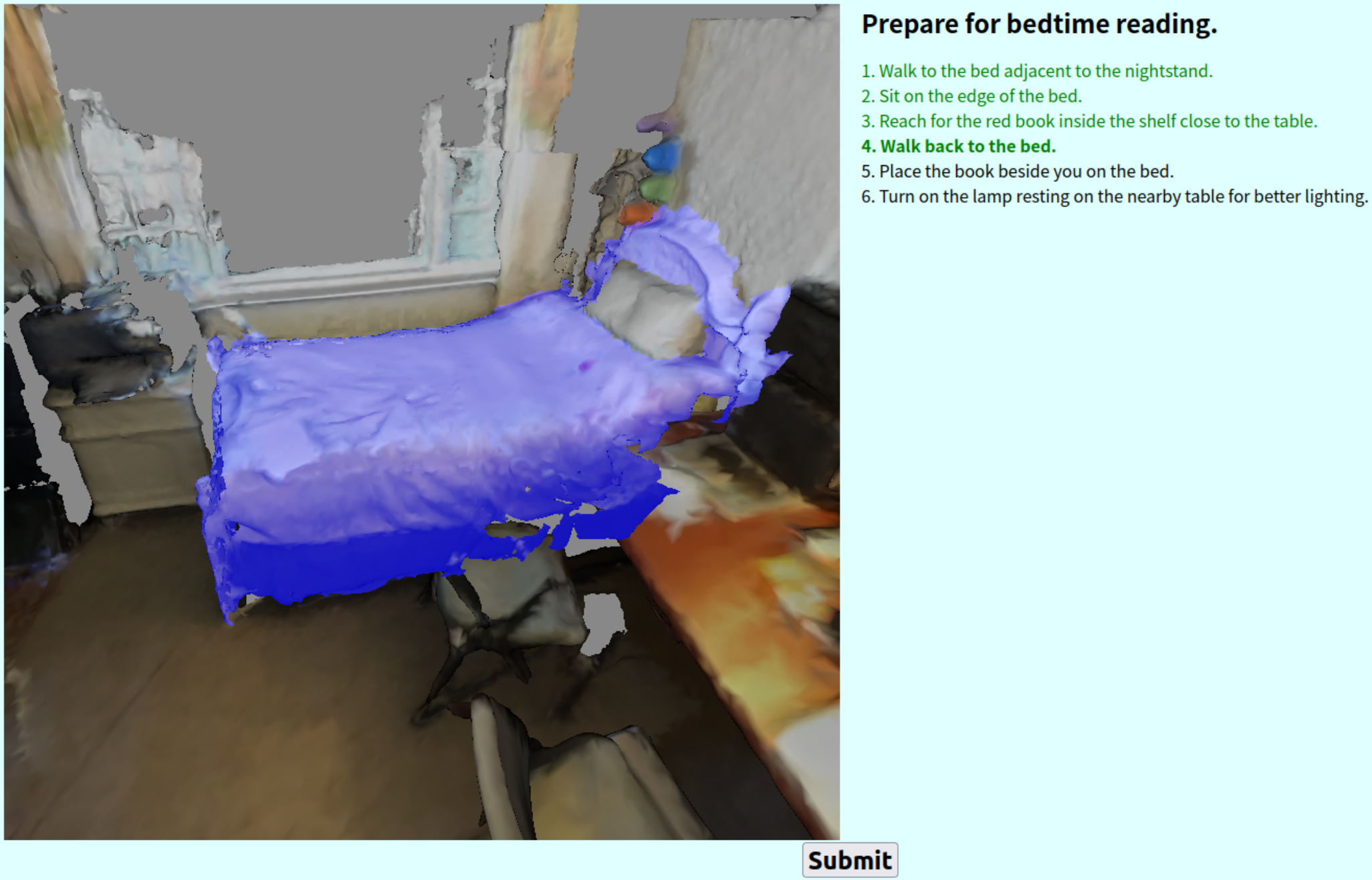}
        \caption{User interface from our human evaluation study, showing a participant selecting the target object for Step 4 of the task.}
        \label{fig:human-study}
\end{figure*}

To evaluate the practical difficulty of our proposed task, we conducted a human study by randomly sampling 100 tasks from the evaluation set of \dataset. Participants were given an interactive 3D scene and a task in a web viewer. Despite some artifacts in the 3D scene viewer, human participants achieved 85\% s-acc and 63\% t-acc, significantly outperforming all evaluated methods. The disparity demonstrates that the \dataset benchmark is indeed challenging for current models. A screenshot of the human study interface to demonstrate the human study workflow is provided in \cref{fig:human-study}.

\begin{figure*}[t]
    \centering
    % \vspace{-5mm}
    \includegraphics[width=\linewidth]{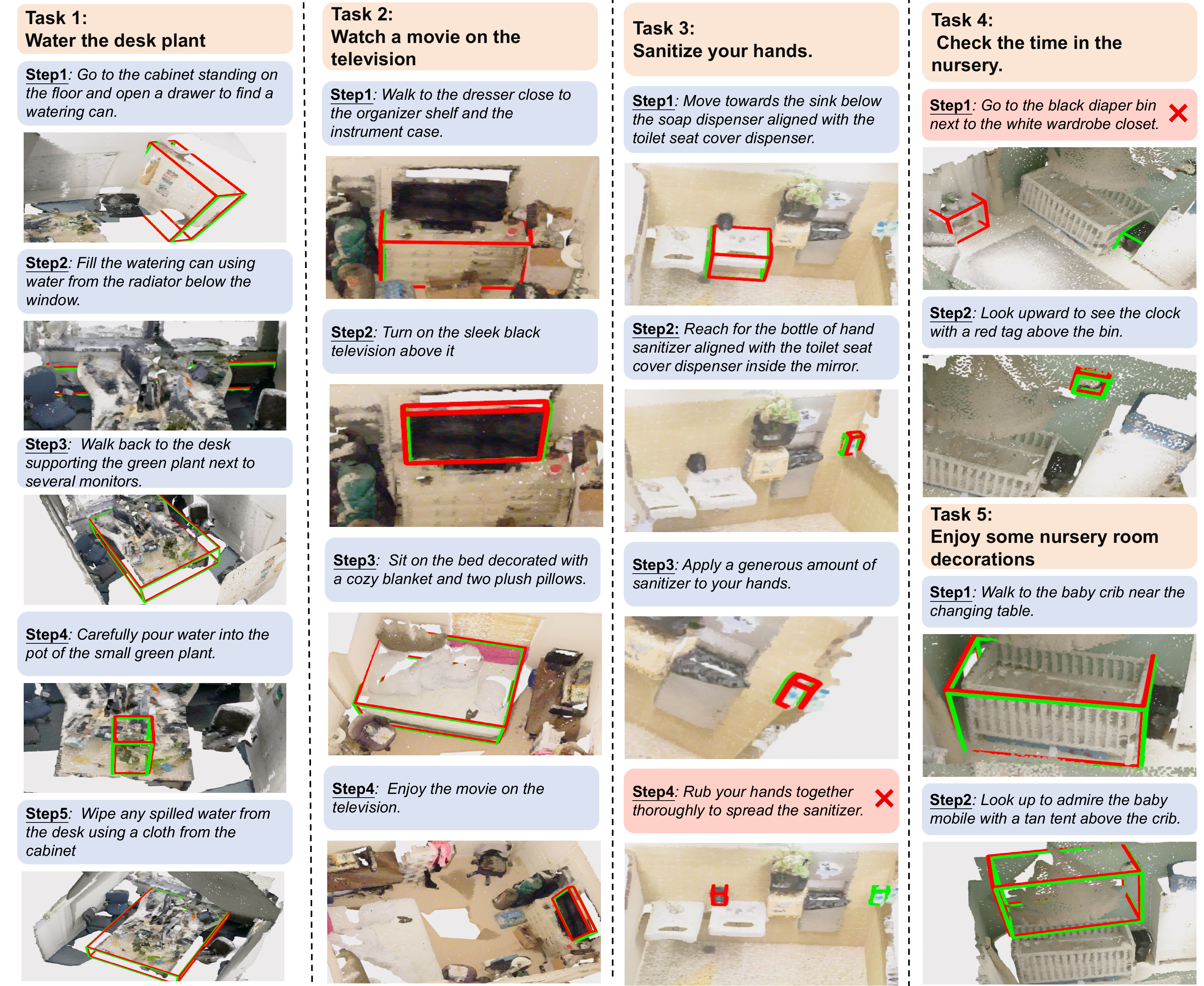}
    \caption{\textbf{Qualitative grounding results from \model.} \textcolor{Red}{Red} are predictions and \textcolor{green}{green} are ground-truth boxes.}
    \label{fig:qualitative}
\end{figure*}

\subsection{Qualitative Grounding Results}
\cref{fig:qualitative} illustrates the critical role of multi-step reasoning in sequential grounding tasks. The results from \model show that after training, the model is capable of performing sequential grounding, as evidenced in tasks 1, 2, and 5. However, the model sometimes struggles to maintain sequential consistency across sequential steps, as observed in task 3. Task 4, in particular, highlights a failure case in which the model fails to grasp the concept of a diaper bin. The examples highlight the challenges and complexities inherent in the sequential grounding task, emphasizing the need for models possessing both robust sequential reasoning abilities and a solid understanding of common-sense knowledge to achieve consistent and accurate results.

% \subsection{Impact on Embodied Navigation}
% While interactive evaluation of action sequences is currently infeasible due to the static nature of the reconstructed 3D scenes, we demonstrate the relevance of our annotations by integrating the LEO model with a navigation module in an embodied setting. Specifically, we use the GreedyGeodesicFollower class from Habitat-Sim~\citep{savva2019habitat} to guide task-oriented navigation within HM3D scenes based on the grounding results (the centers of the target objects). We have provided three navigation videos showcasing this process in the supplementary material.

\subsection{Planning Ability of \model}
In this additional experiment, we evaluate the planning capability of \model fine-tuned on \dataset. Given a task description $t$, \model is required to predict both steps $\{s_1,..,s_n\}$ and target objects $\{o_1,...,o_n\}$, using beam search (width=5) for decoding.

Due to the open-vocabulary nature of step instructions and permissible variations in execution order, exact-match metrics are inadequate for evaluation. Instead, we adopt the GPT-4-based assessment protocol from OpenEQA~\citep{majumdar2024openeqa}, which scores semantic alignment between generated and ground-truth plans on a 1--5 Likert scale (1:irrelevent, 5:perfectly equivalent). \model achieves a GPT-4 evaluation score of $\mathbf{2.1 \pm 1.0}$ on ScanNet, highlighting substantial potential for improvement. Full prompts for GPT-4 scoring are provided in \cref{table::prompt_for_gpt_score}.

\begin{figure*}[th!]
\centering
\begin{minipage}{0.9\linewidth}\vspace{0mm}    \centering
\begin{tcolorbox}
\fontsize{7.0pt}{0.8\baselineskip}\selectfont
You are a helpful assistant that can evaluate the quality of task planning given a scene, a task description, a ground truth task planning, and a predicted task planning.
To mark a response, you should output a single integer between 1 and 5 (including 1, 5), with format \textasciigrave\textasciigrave\textasciigrave Your mark: number\textasciigrave\textasciigrave\textasciigrave.
5 means that the predicted task planning perfectly solves the problem described in the task and matches the ground truth task planning.
1 means that the predicted task planning is completely irrelevant to the task description and does not match the ground truth task planning.\\

The scene is represented by a scene graph in the JSON dictionary format. Each entity in the scene graph denotes an object instance, named `<category>-<ID>'. The `caption' describes the object's attributes, such as `color', `material', etc. The `relations' describes the object's spatial relations with other objects. For example, from the scene graph:\\
\textasciigrave\textasciigrave\textasciigrave\\
{`sofa-1': {`relations': [`to the right of armchair-2', `in front of table-3'], `caption': `Grey velvet sofa with a rectangular shape and a back and arms, suitable for use in a living room.'}, `armchair-2': {`relations': [`to the left of sofa-1'], `caption': `The armchair is made of leather, specifically black leather, and has a spherical shape.'}, `table-3': {`relations': [], `caption': `The table is a rectangular wooden table with a brown finish, sometimes used as a dining table or coffee table, with a smooth wooden texture and various styles, including a sign or place setting on it, and can have plates or a white cloth on it.'}}\\
\textasciigrave\textasciigrave\textasciigrave\\

You can know that `sofa-1' is grey, the `armchair-2' is made of leather, the `table-3' is made of wood, the `armchair-2' is on the left of the `sofa-1', the `sofa-1' is in front of the `table-3'.\\

Using the provided scene graph, you should decide whether predicted task planning can solve the problem described in task description.\\
Here are some examples:\\
\textasciigrave\textasciigrave\textasciigrave\\
<example>\\
\textasciigrave\textasciigrave\textasciigrave\\

Your Turn, output with format \textasciigrave\textasciigrave\textasciigrave Your mark: number\textasciigrave\textasciigrave\textasciigrave.

Scene graph: <scene graph>

Task description: <task description>

Ground truth task planning text: <gt plan text>

Ground truth object id: <gt object id>

Predicted task planning text: <pred plan text>
\end{tcolorbox}
\caption{Prompt messages for computing GPT score.}
\label{table::prompt_for_gpt_score}
\end{minipage}
\end{figure*}

\subsection{Discussions}
\paragraph{Rationale for limiting to one target object per step.} The primary consideration for this restriction is that most mobile manipulators (e.g., the one used in SayCan~\citep{saycan}) are \textit{single-arm} and can manipulate only one object at a time. This design aligns with current practical constraints and facilitates the adaptation of 3D visual grounding models to real-world robotic tasks. Nevertheless, our data generation pipeline is flexible and can be easily adapted to support multi-target actions by adjusting the GPT-4 prompt, as detailed in \cref{fig:prompt messages}.

\paragraph{Handling steps that do not appear to involve a target object.} Some steps, like “Rub your hands” (task 3's step 4 in \cref{fig:qualitative}), involving the agent itself rather than a specific object in the scene, we consider the target object from the previous step as the reference. This implies that no positional change is required, which is a reasonable assumption in the navigation setting. These steps reflect realistic interactions and are part of the task’s natural sequence, so we keep them in our dataset.

% \paragraph{Limitations} The \dataset benchmark adheres to the problem setting of 3D visual grounding based on the assumption that the model can directly access the point cloud representation of the entire scene. This assumption may not hold in real-world applications. Future work will focus on expanding task-oriented sequential grounding within the context of lifelong navigation, requiring the model to engage with the environment through exploratory mechanisms.